\documentclass[conference]{IEEEtran}

\usepackage{times}
\usepackage{epsfig}
\usepackage{graphicx}
\usepackage{cite}
\usepackage{amsmath,amssymb,amsfonts}
\usepackage{algorithmic}
\usepackage{graphicx}
\usepackage{textcomp}
\usepackage{xcolor}

\usepackage[pagebackref=true,breaklinks=true,letterpaper=true,colorlinks,bookmarks=false]{hyperref}

\begin{document}

\title{Streaming Networks: Increase Noise Robustness and Filter Diversity via Hard-wired and Input-induced Sparsity}

\author{\IEEEauthorblockN{Sergey Tarasenko\IEEEauthorrefmark{1},
Fumihiko Takahashi\IEEEauthorrefmark{2}}

\IEEEauthorblockA{Next-Gen Mobility Division, Mobility R$\&$D Group,
JapanTaxi\\
Tokyo, Japan\\
Email: \IEEEauthorrefmark{1}tarasenko.sergey@japantaxi.co.jp,
\IEEEauthorrefmark{2}fumihiko.takahashi@japantaxi.co.jp}}

\maketitle

\begin{abstract}
The CNNs have achieved a state-of-the-art performance in many applications. Recent studies illustrate that CNNs’ recognition accuracy drops drastically if images are noise corrupted. We focus on the problem of robust recognition accuracy of noise-corrupted images. We introduce a novel network architecture called Streaming Networks. Each stream is taking a certain intensity slice of the original image as an input, and stream parameters are trained independently. We use network capacity, hard-wired and input-induced sparsity as the dimensions for experiments. The results indicate that only the presence of both hard-wired and input-induces sparsity enables robust noisy image recognition. Streaming Nets is the only architecture which has both types of sparsity and exhibits higher robustness to noise. Finally, to illustrate increase in filter diversity we illustrate that a distribution of filter weights of the first conv layer gradually approaches uniform distribution as the degree of hard-wired and domain-induced sparsity and capacities increases.
\end{abstract}

\section{Introduction}

\textbf{Brief overview of the CNNs}. Since its first introduction in 1998 by Lecun et al. \cite{LeCun1998GradientbasedLA}, the convolutional neural networks (CNNs) have proved their effectiveness by achieving state-of-the-art solutions for many tasks.

There is a vast variety of CNN architectures in the literature (AlexNet\cite{Krizhevsky2012ImageNetCW}, LeNet \cite{LeCun1998GradientbasedLA}, ResNet\cite{He2015DeepRL}, GoogLeNet \cite{Szegedy2014GoingDW}, VGG \cite{Simonyan2014VeryDC} etc). These are the networks that have a single stream structure.

Recently, CNNs with more than one processing streams have started to gain popularity. To our knowledge, the first two-stream network was introduced by Chorpa \cite{Chopra2005LearningAS} and it is widely known as a ``Siamese network". The motivation behind two streams is that each of the streams carries information about a dedicated image. Images fed to the streams are different.

Most recently, two-stream networks have been used for the vast variety of recognition, segmentation and classification tasks such as similarity assessment (Siamese networks and pseudo-Siamese \cite{Chopra2005LearningAS}\cite{Zagoruyko2015LearningTC}), change detection and classification \cite{Varghese2018ChangeNetAD}, action recognition in videos \cite{Simonyan2014TwoStreamCN}, one-shot image recognition \cite{Koch2015SiameseNN}, simultaneous detection and segmentation \cite{Hariharan2014SimultaneousDA}, human-object interaction recognition \cite{Gkioxari2017DetectingAR}, group activity recognition \cite{Azar2018AMC}, etc.

\textbf{The CNNs and the primate brain.} Regarding the signal propagation in the brain networks, Thorpe at el. \cite{Thorpe1996SpeedOP,Thorpe2001SpikebasedSF,VanRullen2002SurfingAS} argued that the stronger the response of a given neuron, the faster such response should be produced, meaning that it takes less time to produce stronger output than a weaker one. 

Thorpe et al. have suggested that neural outputs produced nearly at the same time form \textit{waves of spikes}. So even information about the static single image is propagated through the neural network in time separated packets called waves of spikes, thus a static image is unfolded in time due to different response time for stronger and weaker outputs.

The idea of waves of spikes was then employed by Tarasenko \cite{TarasenkoWaves2011}. Tarasenko continued work by Serre et al. \cite{Serre2007RobustOR} by proposing an on-line learning method for feature extraction and extending the pseudo-CNN to implement a predictive coding \cite{Rao1999PredictiveCI} mechanism. 

\begin{figure}[t]
\begin{center}
\includegraphics[width=0.8\linewidth]{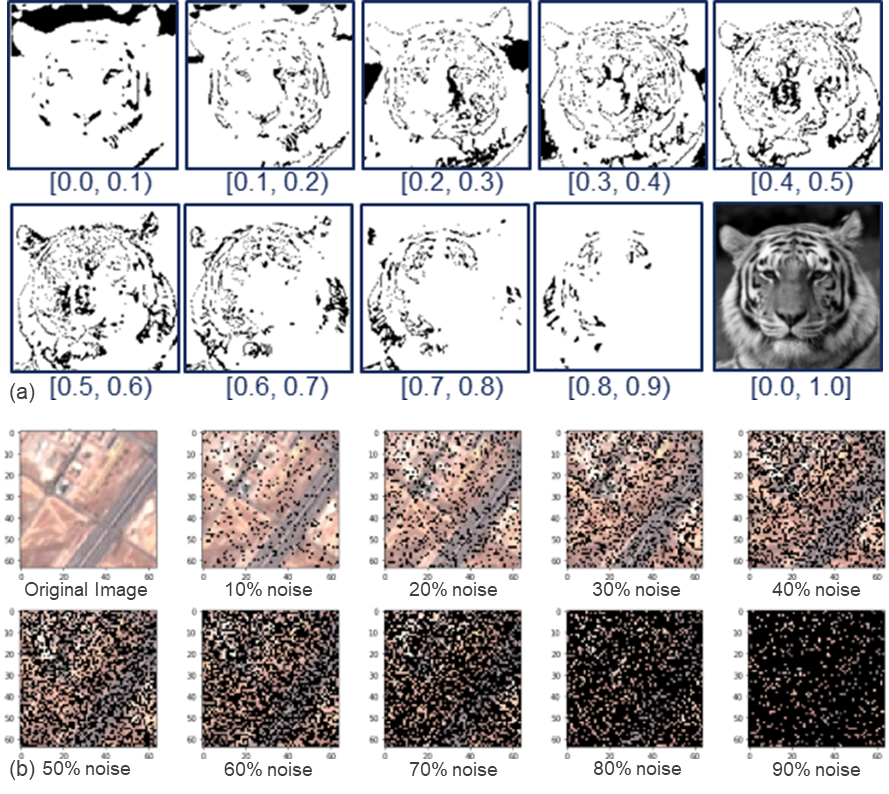}
\end{center}
\caption{Intesity slices and Noise.}
\label{fig:noisy}
\end{figure}

The important peculiarity of work by Tarasenko is that to extract features images, containing complete information were used, while to enable mechanism of predictive coding after feature extraction, images were fed into the network by intensity slices (similar to waves of spikes). Examples of image intensity slices are presented in Fig. \ref{fig:noisy}(a). Every single image with normalized pixel values was split into 10 images, which correspond to one of the intensity slices ranging from 0.0 to 1.0 with step 0.1. Then these slices were consecutively propagated into the network.

\section{Related Work}
\textbf{The CNNs and image distortions.} Recent studies have illustrated that CNNs' performance is extremely fragile for distortions of the input images such as noise, image occlusions, rotation, scaling, etc. \cite{Szegedy2013IntriguingPO}\cite{Eykholt2017RobustPA}. In this paper, we approach the issue of robust recognition when images are corrupted with random zero-noise, i.e., a certain portion of pixels across the entire image is randomly set to zero intensity value.

The topic of robust recognition by CNNs under conditions of noise has been explored in \cite{Aghdam2016AnalyzingTS,Aghdam2016IncreasingTS}.

In work \cite{Aghdam2016AnalyzingTS}, authors analyzed the robustness (stability) of CNNs against image degradation due to noise. The same group of authors suggests a method to increase the stability of CNNs by introducing a denoising layers \cite{Aghdam2016IncreasingTS}.

\textbf{Sparsity.} 
The networks in the brain exhibit very high sparsity degree \cite{Ranzato2007SparseFL} . The topic of sparsity is well establashed in studies by \cite{Ranzato2007SparseFL} . There are two approahes to sparsity: 1) sparsity of representation \cite{Olshausen1997SparseCW}\cite{Olshausen1996EmergenceOS} and 2) for network training \cite{SalakhutdinovCode}, \cite{Masters2015vol1}.

\begin{figure}[t]
\begin{center}
\includegraphics[width=0.7\linewidth]{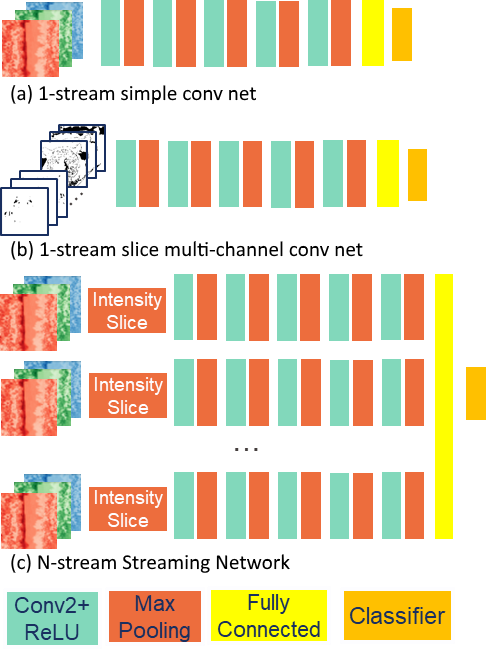}
\end{center}
\caption{One stream CNN and Streaming Network architectures.}
\label{fig:stnet}
\end{figure}

\section{Streaming Networks}

\textbf{General Definition} 
Here we introduce a novel CNN architecture. We take a 1-stream CNN and add image intensity slicing module as an input. Then we clone new networks with intensity slices set to extract different intensity slices. Each such network constitutes a single stream. Finally, we concatenate outputs of all the streams to one fully connected layer, which is connected with a classifier. A number of fully connected layers after concatenated layer can vary. 

The weights and biases within every single stream are decoupled from the ones in other streams and are trained independently.

We call this architecture a Streaming Network (Fig. \ref{fig:stnet}(c)). 

\textbf{Streaming Networks as an Orthogonal Basis for Image Representation}

\begin{figure}[t]
\begin{center}
\includegraphics[width=1.0\linewidth]{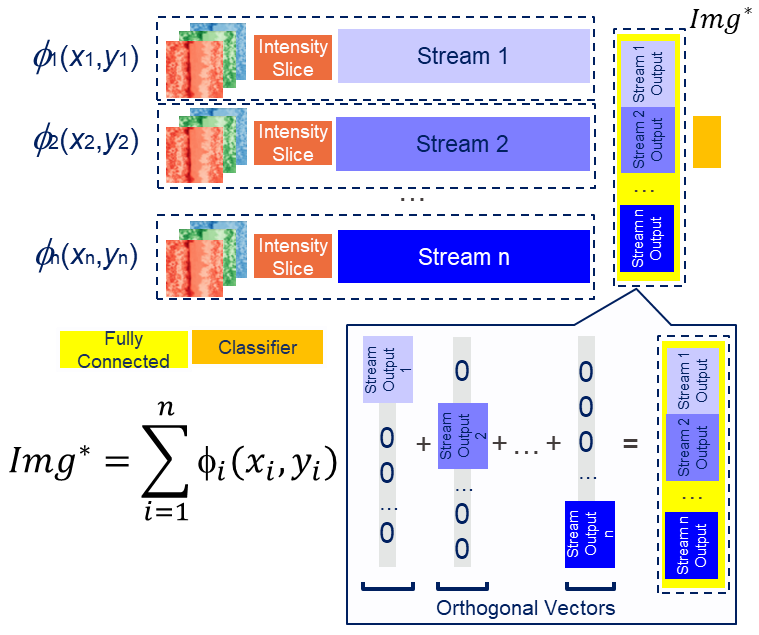}
\end{center}
\caption{Streaming Network as an orthogonal basis for image represantation.}
\label{fig:stream_basis_unit_weights}
\end{figure}

In this section, we explain the mathematical concepts behind Streaming Networks. Eq. (\ref{eq:basis_vectors}) introduces an image representation by means of the orthogonal vectors, which form an orthogonal basis of some image representation space:
\begin{equation}
Img^{*} = \sum^{n}_{i=1} \alpha_{i} v_{i}
\label{eq:basis_vectors}
\end{equation}

where $\alpha_{i}$ is a weight coeffient for $i$-th orthogonal basis vector $v_{i}$, $i$ = 1, ..., $n$, and $Img^*$ is an image representation.

As the next step, we can generalize eq. (\ref{eq:basis_vectors}) by providing orthogonal basis functions $\phi_{i}(x_{i},y_{i}) $ instead of basis vectors $v_{i}$ \cite{Olshausen1996EmergenceOS}:

\begin{equation}
Img^{*} = \sum^{n}_{i=1} \alpha_{i}\phi_{i}(x_{i},y_{i})
\label{eq:basis_functions}
\end{equation}

where pair ($x_{i},y_{i}$), $i$ = 1, ..., $n$, represents some part of an original image.

If now, we set all weighting coefficients $\alpha_i$, $i$=1,...$n$, to 1, we will get eq. (\ref{eq:basis_functions_unit_weight}):

\begin{equation}
Img^{*} = \sum^{n}_{i=1} \phi_{i}(x_{i},y_{i})
\label{eq:basis_functions_unit_weight}
\end{equation}

Eq. (\ref{eq:basis_functions_unit_weight}) perfectly describes the Streaming Network architecture (Fig. \ref{fig:stream_basis_unit_weights}): functions $\phi_{i}(\cdot)$, $i$=1,...$n$, correspond to the streams of a Streaming Network, pairs ($x_{i},y_{i}$), $i$=1,...$n$, represent some transformation of parts of the original image, e.g., intensity slices.

\begin{figure}[t]
\begin{center}
\includegraphics[width=1.0\linewidth]{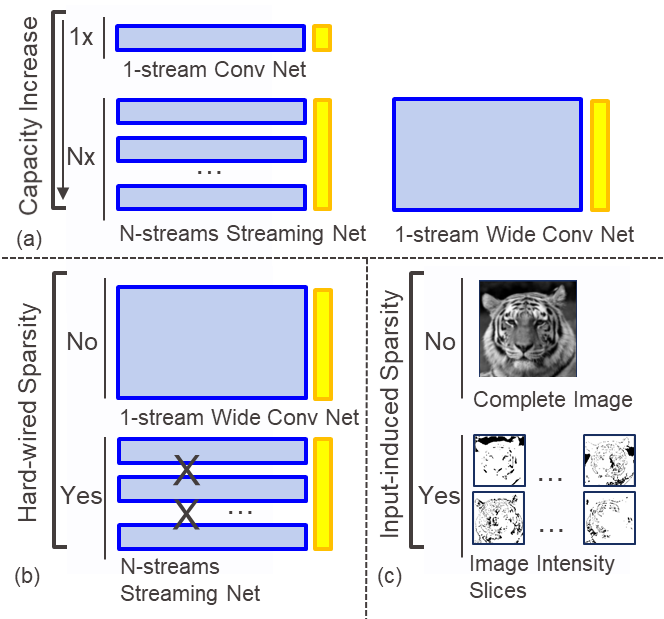}
\end{center}
\caption{Experimental dimensions explained: (a) network capacity increase; (b) hard-wired sparsity; (c) input-induced sparsity.}
\label{fig:exper_dims}
\end{figure}

\begin{figure}[t]
\begin{center}
\includegraphics[width=1.0\linewidth]{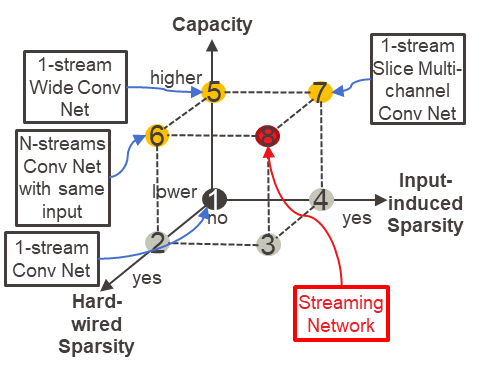}
\end{center}
\caption{Experimental design with corresponding network achitectures.}
\label{fig:exper_design}
\end{figure}

\begin{figure*}[t]
\begin{center}
\includegraphics[width=1.0\linewidth]{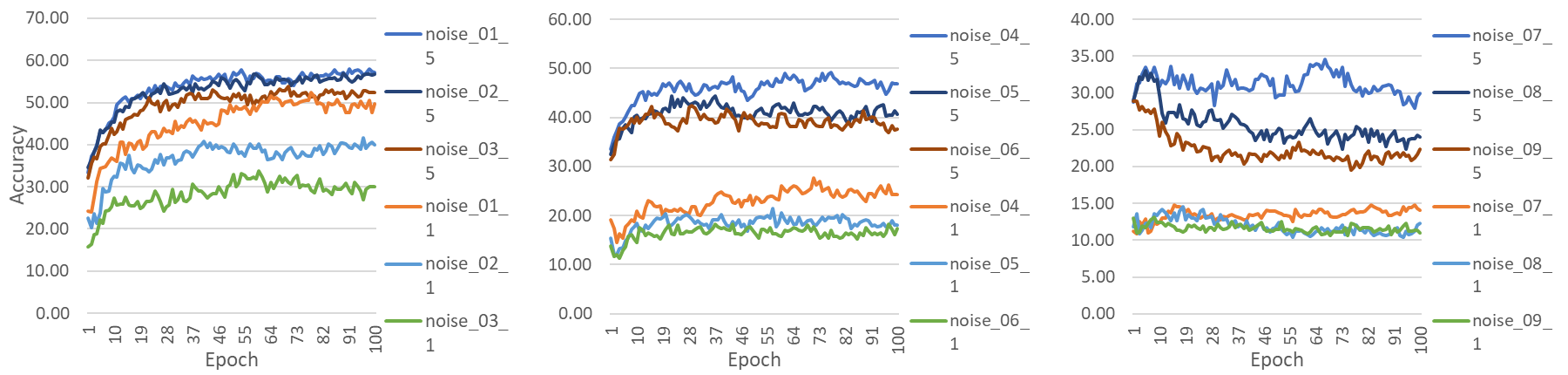}
\end{center}
\caption{Tests of Cifar10 dataset using Adam optimizer with 0.0001 learning rate for 1-stream simple CNN vs. 5-stream Streaming Net. The legend codes experimental conditions as noise$\_$$\{$ noise-ration$\}$$\_$$\{$network$\}$. For example, noise$\_$01$\_$1 implies 1-stream simple network with noise 10$\%$. By default we consider that networks with more than one stream are Streaming Networks. All curves are averaged across multiple runs.}
\label{fig:res_cifar10_0001}
\end{figure*}

\begin{figure*}[t]
\begin{center}
\includegraphics[width=1.0\linewidth]{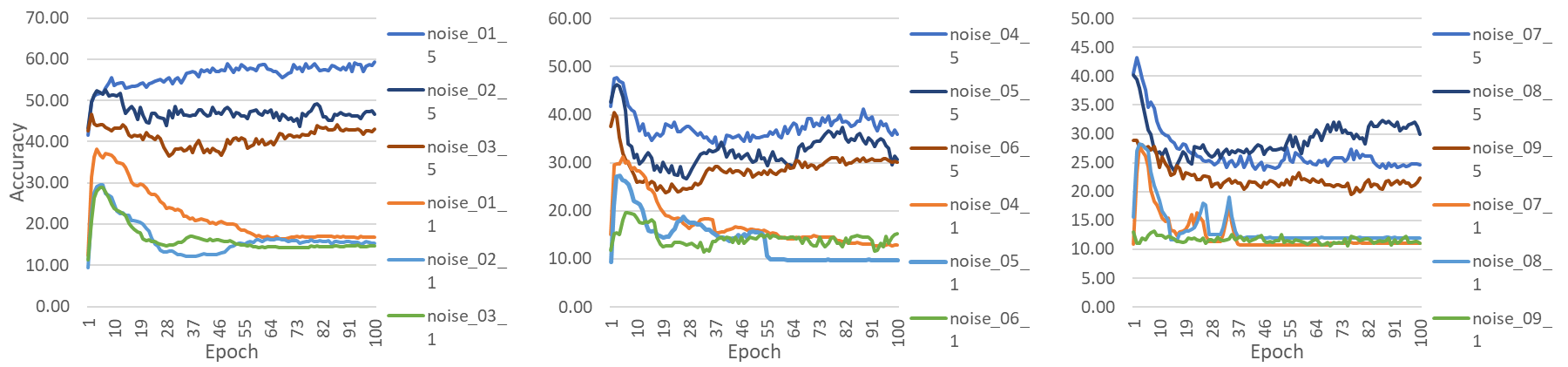}
\end{center}
\caption{Tests of Eurosat dataset using Adam optimizer with 0.0001 learning rate for 1-stream simple CNN vs. 5-stream Streaming Net. The notations are the same as in Fig. \ref{fig:res_cifar10_0001}.}
\label{fig:res_euro_0001}
\end{figure*}

\begin{figure*}[t]
\begin{center}
\includegraphics[width=1.0\linewidth]{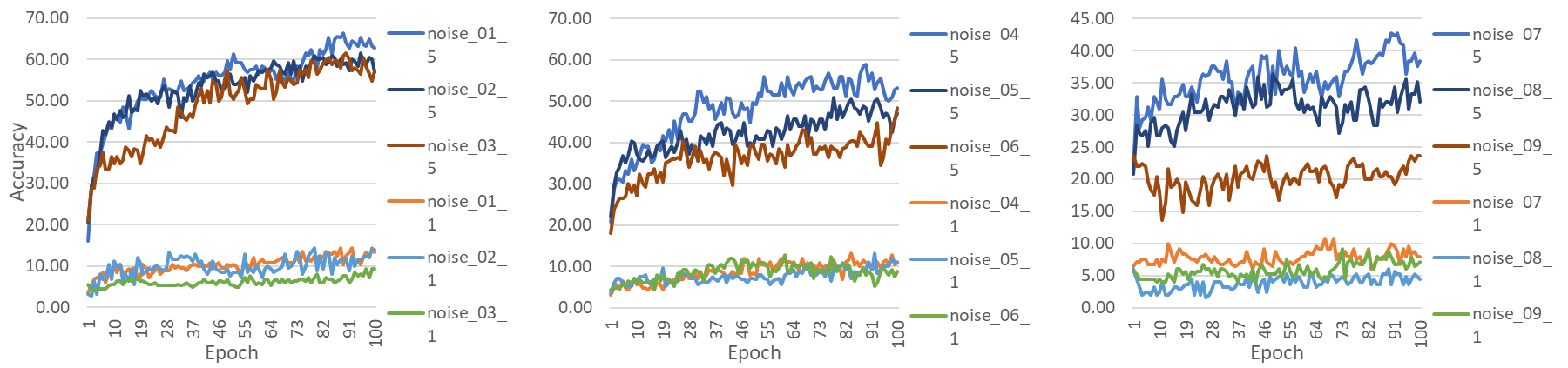}
\end{center}
\caption{Tests of UCmerced dataset using Adam optimizer with 0.0001 learning rate for 1-stream CNN and 5-stream Streaming Network. The notations are the same as in Fig. \ref{fig:res_cifar10_0001}.}
\label{fig:res_ucmerced_0001}
\end{figure*}

\begin{figure*}[t]
\begin{center}
\includegraphics[width=1.0\linewidth]{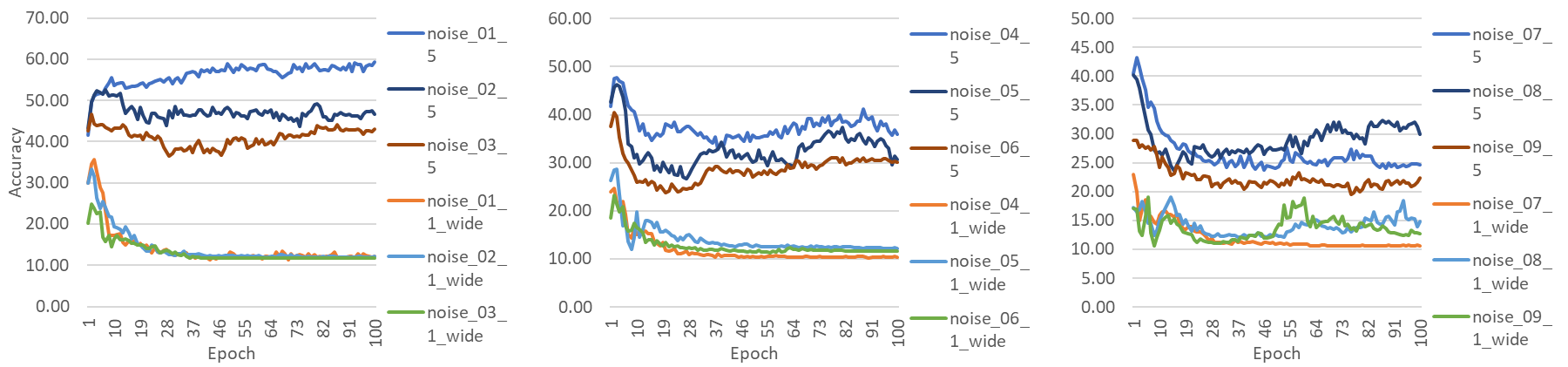}\end{center}
\caption{Tests of Eurosat dataset using Adam optimizer with 0.0001 learning rate for 1-stream wide CNN vs. 5-stream Streaming Net. The notations are the same as in Fig. \ref{fig:res_cifar10_0001}.}
\label{fig:res_euro_1stream_wide}
\end{figure*}

\begin{figure*}[t]
\begin{center}
\includegraphics[width=1.0\linewidth]{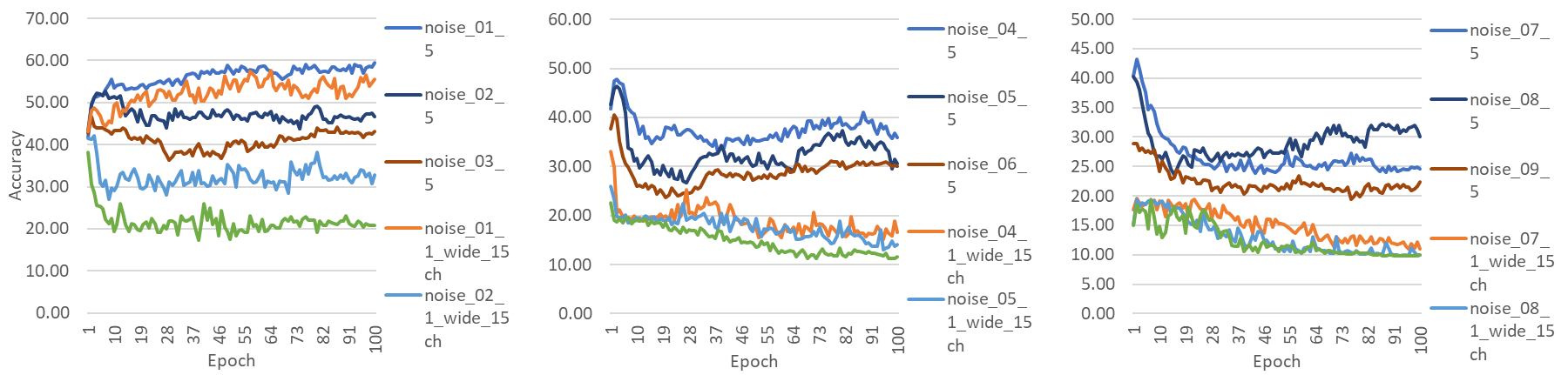}\end{center}
\caption{Tests of Eurosat dataset using Adam optimizer with 0.0001 learning rate for 1-streams 15-channel wide vs. 5-stream Streaming Net. The notations are the same as in Fig. \ref{fig:res_cifar10_0001}.}
\label{fig:res_euro_15ch_5_0001}
\end{figure*}

\begin{figure*}[t]
\begin{center}
\includegraphics[width=1.0\linewidth]{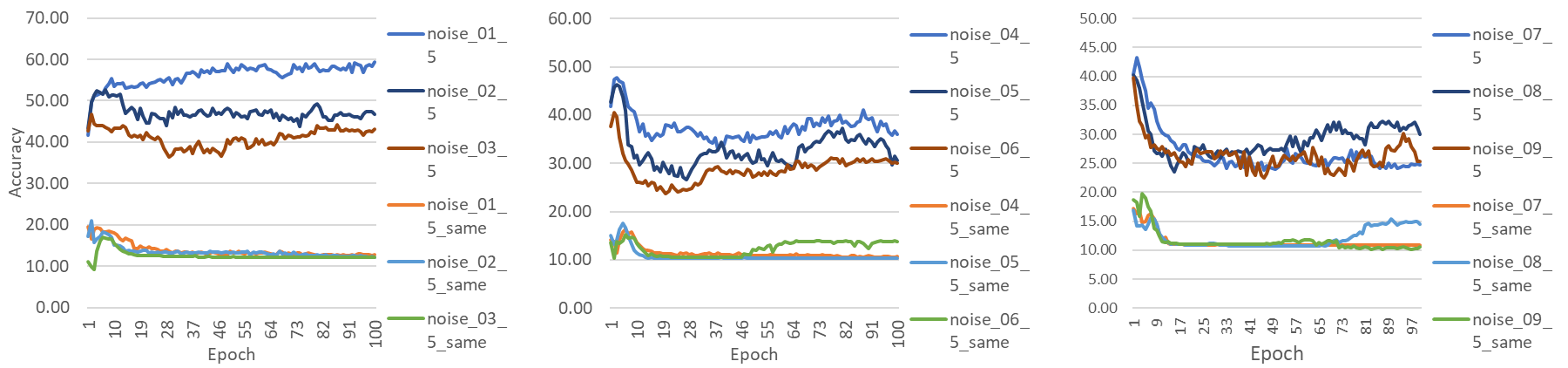}\end{center}
\caption{Tests of Eurosat dataset using Adam optimizer with 0.0001 learning rate for 5-streams with the same input vs. 5-stream Streaming Net. The notations are the same as in Fig. \ref{fig:res_cifar10_0001}.}
\label{fig:res_euro_5same}
\end{figure*}


\section{Experiments}

In the previous section, we have outlined the issue of accuracy reduction when images are corrupted in some way and touched the topic of spacity, which is expected to produce robust encoding. In this section, we would like to invetigate if it is possible to increase robustness of the CNNs introducing spasity dimension.

The gist of sparsity control for network training is to make neurons to be activated with a certain frequency. Thus, providing a diversity of paths within a neural network. The concept of sparsity control makes a perfect sense in terms of binary neurons, which produce either 1, when active, or 0, when in-active. If neuron always produces 1 or 0, it is useless from the point of view of classification. To reduce the number of such neurons, usually, additional term, which penalized the loss function whenever activation frequency of neurons is deviating from the target frequency (usually set to 20$\%$), is introduced. As mentioned above, the concept of sparsity makes perfect sense in terms of binary neurons. It can be extended to sigmoid or tahn neurons by introducing sum average value for activations, however, it is becoming very difficult to access in terms of CNNs with ReLUs, since the output of neurons is not limited on the interval (0,+$\infty$).

To overcome the problem of sparsity control in CNNs with ReLU, we introduce two new concepts regarding sparsity in section Experimental Design \ref{exper_design}.

\subsection{Experimental Design}
\label{exper_design}
In this study, we introduce two approaches to control the sparsity: 1) \textit{hard-wired sparsity}, when we eliminate all-to-all connections in the network by introducing independent streams; 2) \textit{input-induced sparsity}, when we introduce a unique intensity slice of an image to each stream, which refrains independent streams from learning from the same input.

We also add the dimension of network $capacity$, i.e., a total number of filters inside each convolution layer of a given network.

Next we use these three dimensions to construct neural network achitectures (Fig. \ref{fig:exper_dims}). In this study, we qualitative evaluate capacities as $lower$ or $higher$. The hard-wired and input-induced sparsities are considered to be either employed ($yes$ value) or not employed ($no$ values). Using this binary values of dimensions, it is possible to introduce our experimental space in the form of a cube in $Capacity$ - \textit{Hard-wired sparsity} - \textit{Input-induced sparsity} 3D space (Fig \ref{fig:exper_design}).

Each vertex of the cube is related to a particular set of values for each dimension and has a corresponding neural network architecture.

Vertex (1) is charactarize with $lower$ capacity of the network, $no$ hard-wired sparsity and $no$ input-induced sparsity. The network architecture, which corresponds to vertex (1) is \textit{a 1-stream simple CNN} (Fig. \ref{fig:stnet}(a)).

A 1-stream simple CNN has the following structure: 1) conv layer with 32 7x7 filters plus ReLU activation and 2x2 Max-pooling layers;
2) conv layer with 64 5x5 filters plus ReLU activation and 2x2 Max-pooling layers;
3) conv layer with 128 3x3 filters plus ReLU activation and 2x2 Max-pooling layers;
4) conv layer with 256 1x1 filters plus ReLU activation and 2x2 Max-pooling layers;
5) conv layer with 4\footnote{For cifar10 dataset we use 10 1x1 filters} 1x1 filters plus ReLU activation and 2x2 Max-pooling layers;
6) fully connected layer;
7) SoftMax layer with the number of output neurons corresponding to the number of classes.

In our experiments, we consider network architectures with higher capacity, while we take 1-stream simple CNN as a bench-mark. Furthermore, vertices (2), (3) and (4) are lower capacity analogs of vertices (5), (6), (7) and (8). Therefore, we exclude vertices (2), (3) and (4).

Hereafter, we introduce each vertex in the form of 3D vector $\{$$Capacity$ - \textit{Hard-wired sparsity} - \textit{Input-induced sparsity}$\}$. Thus, vertex (1) is coded as $\{$lower, no, no$\}$.

Vertex (5) is coded as $\{$higher, no, no$\}$ and the corresponding network architecture is \textit{a 1-stream wide CNN}, which has $Nx$ filters in each conv layer, where is $N$ an integer multiplier and $x$ is a number of filters in corresponding conv layers of the 1-stream simple CNN. Vertex (6) is coded as $\{$higher, yes, no$\}$ and the corresponding network architecture is \textit{N-stream CNN with the same input for each stream}. Vertex (7) is coded as $\{$higher, no, yes$\}$. We assume that the corresponding architecture is a 1-stream wide network, where image slices are packed into different input channels. Thus, if we cut an original RGB image with three channels into five intensity slices, then input is combined from 15 channels - 3 channels per a single slice. We call such network architecture \textit{1-stream slice multi-channel CNN}. We assume that \textit{a 1-stream slice multi-channel wide CNN} (Fig. \ref{fig:stnet}(b)) corresponds to vertex (7). Finally, vertex (8) is coded as $\{$higher, yes, yes$\}$ and the corresponding network architecture is a \textit{Streaming Network}.

\subsection{Datasets and Learning Procedure}
For our experiments we use three datasets. The selected datasets are cifar10\footnote{https://www.cs.toronto.edu/~kriz/cifar.html}, Eurosat (rgb)\footnote{https://github.com/phelber/eurosat} \cite{Helber2017EuroSAT} and UCmerced land use\footnote{http://weegee.vision.ucmerced.edu/datasets/landuse.html}.

The reason to select these three dataset is to test our approach on different types of objects. Cifar10 dataset contains RGB 32x32 images of 10 classes (airplane, automobile, bird, cat, deer, dog, frog,horse, ship, truck), while Eurosat and UCmerced consist of 10 and 21 classes, respectively, of aerial satelite images. Thus, selecting these datasets we test images of various origins.

For all our experiments, we use Adam optimizer with learning rate of 0.0001 accompanied by $\beta$1 = 0.99, $\beta$2 = 0.9 and $\epsilon$ = 1e-08, and run all the trainings for 100 epochs. Throughout the experiments, we use SoftMax classifier.

For each dataset we run the networks for noise level (ratio of pixels corrupted with noise) ranging from 0.1 to 0.9 with step 0.1, thus constituting 9 different levels. Examples of different noise levels for a selected Eurosat image are illustrated in Fig.\ref{fig:noisy}.

When we train the network after each iteration we compute network accuracy for test data without noise and the test data corrupted with noise. In the figure, however, we only illustrates results for noise-corrupted images.

\subsection{Objective}
Our objective is to investigate if it is possible to increase CNN robustness to noise by introducing two types of sparcity, i.e., input-induced and hard-wired. We do this by measuring recognition accuracy for noisy images for each type of five network architectures.

\begin{figure*}[t]
\begin{center}
\includegraphics[width=1.0\linewidth]{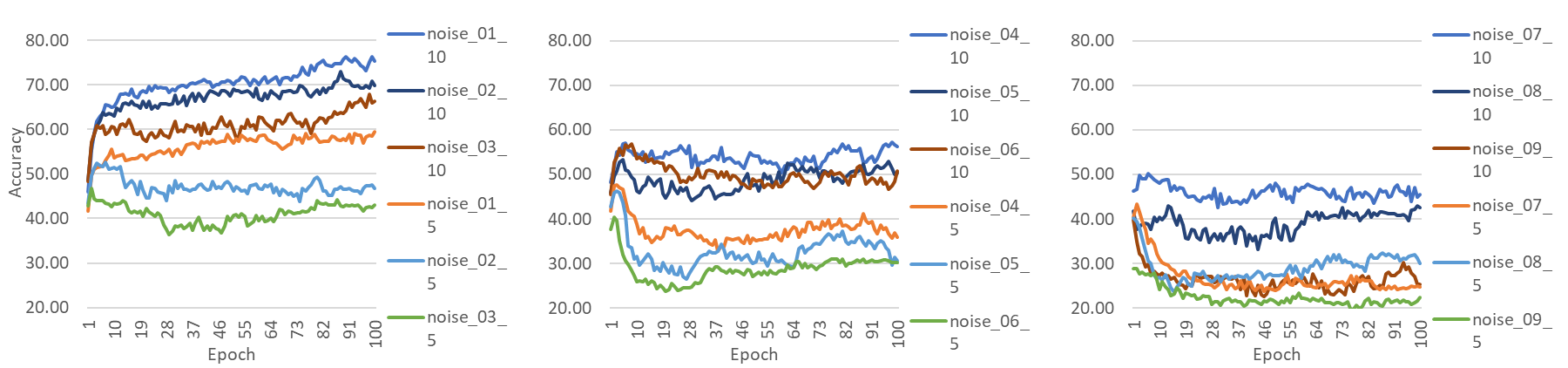}
\end{center}
\caption{Tests of Eurosat dataset using Adam optimizer with 0.0001 learning rate for 5-streams vs. 10-stream Streaming Net. The notations is the same as in ig. \ref{fig:res_cifar10_0001}.}
\label{fig:res_euro_5x10_0001}
\end{figure*}

\begin{figure*}[t]
\begin{center}
\includegraphics[width=1.0\linewidth]{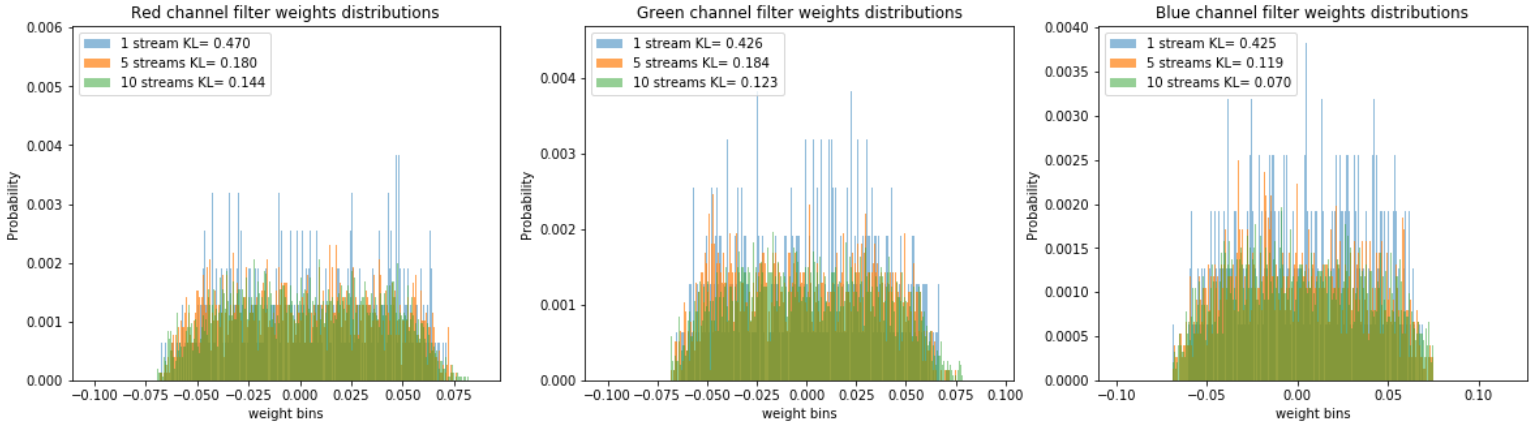}
\end{center}
\caption{Distribution of weights Eurosat dataset with 0.0001 learning rate.}
\label{fig:euro_weight_distrib_all_streams}
\end{figure*}

\begin{figure*}[t]
\begin{center}
\includegraphics[width=1.0\linewidth]{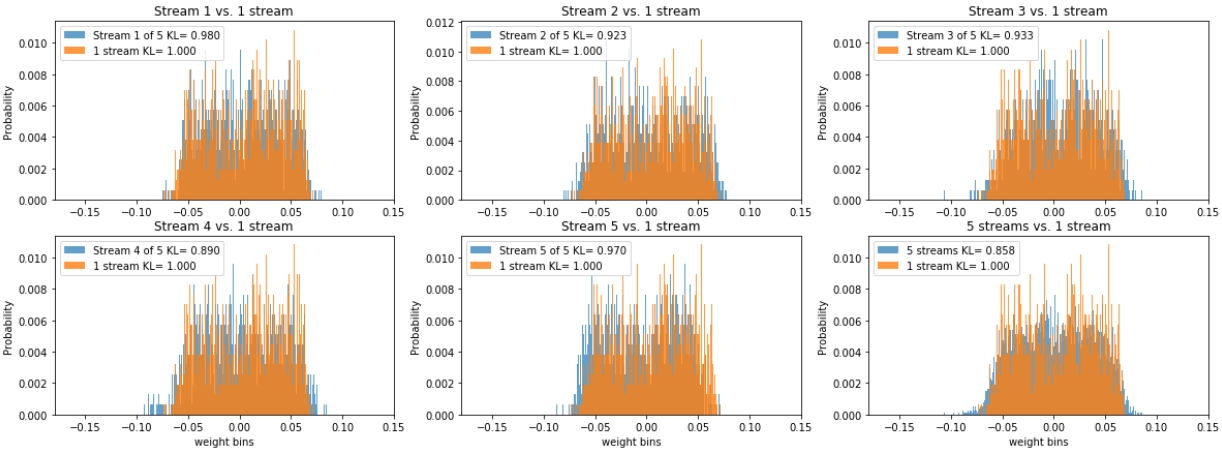}
\end{center}
\caption{Distribution of filter values across all layers for each stream obtain for Eurosat dataset with 0.0001 learning rate.}
\label{fig:euro_weight_distrib_5streams}
\end{figure*}

\begin{figure*}[t]
\begin{center}
\includegraphics[width=1.0\linewidth]{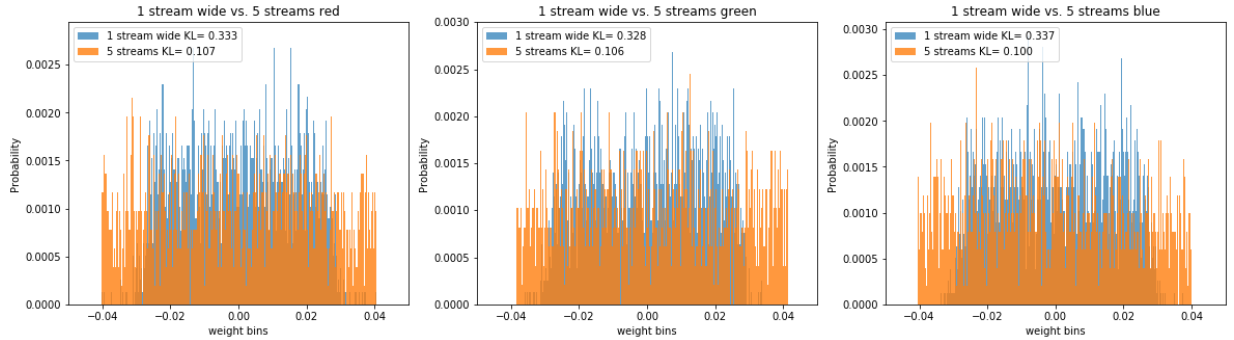}
\end{center}
\caption{5 streams Streaming Network weights vs. 1 stream wide CNN.}
\label{fig:5_wide}
\end{figure*}

\begin{figure*}[t]
\begin{center}
\includegraphics[width=1.0\linewidth]{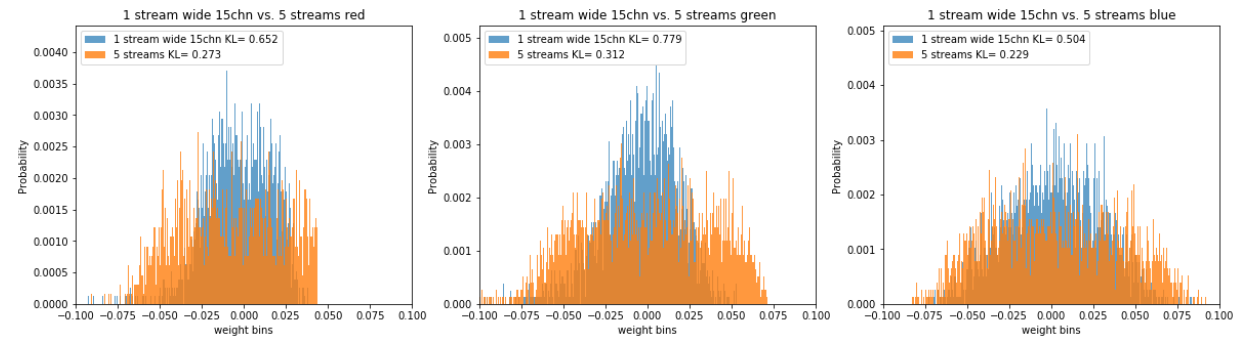}
\end{center}
\caption{5 streams Streaming Network weights vs. 1 stream wide 15-channel CNN.}
\label{fig:5_wide_15chn}
\end{figure*}

\begin{figure*}[t]
\begin{center}
\includegraphics[width=1.0\linewidth]{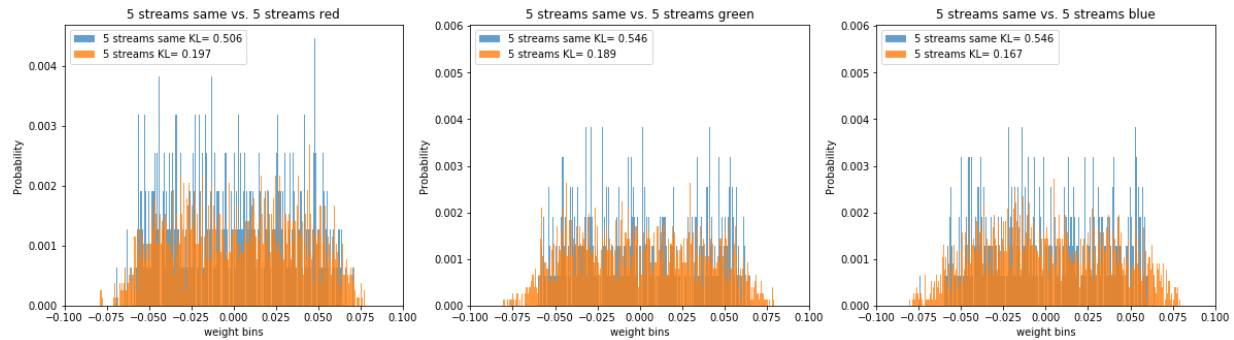}
\end{center}
\caption{5 streams Streaming Network weights vs. 5 streams same input CNN.}
\label{fig:5_5streams_same}
\end{figure*}

\subsection{Results}

\textbf{Streaming Net vs. 1-stream simple CNN}. For all the dataset we computed accuracy for noise-corrupted images. The results for performance comparison of 1-stream simple CNN vs. 5-streams Streaming Net are presented in Figs. \ref{fig:res_cifar10_0001}, \ref{fig:res_euro_0001} and \ref{fig:res_ucmerced_0001} for cifar10, Eurosat and UCmerced datasets, respectively. These figures indicate that a 5-streams Streaming Net outperforms 1-stream simple CNN in recognition accuracy for noisy images.

\textbf{Streaming Net vs. 1-stream wide CNN}. Hereafter, we focus on the results for Eurosat datasets to illustrate performance networks, corresponding to vertices (5), (6), (7) and (8). The same results for cifar10 and UCmerced datasets are presented in Supplementary material.

Next, we compare 5-streams Streaming Netw vs. 1-stream wide CNN. Fig. \ref{fig:res_euro_1stream_wide} indicates that the recognition accuracy of noise images degrades dramatically as noise ratio is increasing for 1-stream wide CNN, while it shows high robustness in the case of 5-streams Streaming Net.

\textbf{Streaming Net vs. 1-stream wide slice multi-channel CNN}. In Fig. \ref{fig:res_euro_15ch_5_0001} recognition accuracy for 1-stream slice 15-channel CNN is presented vs. performance of 5-stream Streaming Net. In terms of robustness to noise, a 1-stream slice 15-channel CNN is somewhat more stable to low levels of noise (up 30$\%$) when 1-stream simple CNN, however, it shows no robustness for higher levels of noise. A 5-streams Streaming Network outperforms a 1-stream wide slice 15-channel CNN in noisy images recognition accuracy.

\textbf{Streaming Net vs. 5-stream CNN with same input}. A 5-streaming CNN with the same input is a very similar architecture to Streaming Net as it has multiple streams. However, the key difference is that Streaming Net takes different inputs into streams. Fig. \ref{fig:res_euro_5same} illustrates that noisy images recognition accuracy for a 5-stream CNN with the same inputs is dramatically worse than the one for a 5-streams Streaming Net.

Based on the comparison of neural network architectures for vertices (1), (5), (6), (7) and (8), the intermediate conclusion is that the Streaming Net is the only architectures robust to noise.

\textbf{Streaming Net with more streams}. Next, we investigate if it is possible to further improve noisy images recognition accuracy. We consider a 10-streams Streaming Network with intensity slices [0.0,0.1), [0.1,0.2), [0.2,0.3), [0.3,0.4), [0.4,0.5), [0.5,0.6), [0.6,0.7), [0.7,0.8), [0.8,0.9) and [0.9,1.1).

We compare performance of 10-stream Streaming Net vs. 1-stream wides networks and 5-stream Streaming Net. Fig. \ref{fig:res_euro_5x10_0001} illustrates that a 10-streams Streaming Net provides even higher robustness to noise compared to a 5-streams Streaming Net. 

\begin{table*}
\begin{center}
\begin{tabular}{|l|c|c|c||c|c|}
\hline
Network Architecture & Capacity & Hard-wired & Input-induced & No Noise & Noise \\
& & Spasity & Sparsity & Accuracy Up & Robustness \\
\hline\hline
(1) & Lower & No & No & N/A & No \\

(5) & Higher & No & No & Yes & No \\


(6) & Higher & Yes & No & Yes & No \\


(7) & Higher & No & Yes & Yes & No \\

(8) \textbf{Streaming Network} & Higher & Yes & Yes & Yes & Yes \\
\hline
\end{tabular}
\end{center}
\caption{Experimental results: comparison of the network architectures.}
\label{tab:results}
\end{table*}

\subsection{Distribution of Trained Filter Values}

In this section, we investigate distributions of values in the trained filters of the first conv layers. We perform our analysis from two perspectives: 1) distribution of filter weights\footnote{we consider values within filters as weights. To distinguish the weights within filters from weights between fully connect layer and classifier layer, we call filter values to be the filter weights.} within a single convolution layer; 2) distribution of filter weights first conv layers for all streams.

Our hypothesize is based on the study by Lecun et al. \cite{LeCuneffprop98}, which implies that higher filter diversity enables faster learning and convergence. We hypothesis that emergence of the noise robustness, which we have observed in the experiments about originates from the higher diversity of filters produced by the Streaming Nets.

We assume that the filter weights are sampled from some restricted segment since there can be no infinitely big and infinitely small weights.

Since we sample filter weights from the segment, the most diversity of filter weights is expected from the distribution with the highest entropy. According to Bening and Korolev\cite{beningKorolev2002}, the uniform distribution has the highest entropy value among all distributions spanned upon the restricted segment. Therefore, we expect uniform distribution to be a limit distribution, to which distribution of all filter weights is asymptotically approaching.

To measure how similar a given distribution is to the discrete uniform distribution, we compute Kullback–Leibler (KL) divergence between the distribution of filter weights and uniform distribution.

\begin{equation}
D_{KL}(p||q) = \sum^{N}_{i=1}p(x_{i}) \cdot log \frac{p(x_{i})}{q(x_{i})}
\label{kld}
\end{equation}

By extending Lecun's results, we proposed two hypotheses: introduction of more streams and slices will result in 1) an increase of filter weights diversity; and 2) that the distribution of all filter weights across all first layers of the Streaming Nets will gradually converge in terms KL divergence to a discrete uniform distribution.

\textbf{Distribution of the filter weights across individual conv layers}.
Here we discuss distributions of filter weights for 1-stream simple CNN, 5- and 10-streams Streaming networks in the case of Eurosat data set. In Fig. \ref{fig:euro_weight_distrib_all_streams}, we illustrate the distribution of filter weight for the first conv layer of the 1-stream CNN and 5- and 10-stream Streaming Network separately for each color channel. As we hypothesized above, the KL divergence value decreases as we increase the number of channels. Therefore, the distribution of filter weights is gradually approaching discrete uniform distribution. The distribution for 10-stream Streaming Net is extremely close to a uniform distribution in Fig. \ref{fig:euro_weight_distrib_all_streams}. The weight distribution is derived from single runs of the network. In total, we have analyzed eight triplets of networks. The figures for remaining triplets are presented in Supplementary materials.

Furthermore, we compare the filter disctributions accross all fisrt conv layers for each color channel with the corresponding distributions for a 1-stream wide CNN (Fig.\ref{fig:5_wide}), a 1-stream wide slice 15-channel CNN (Fig.\ref{fig:5_wide_15chn}) and a 5-streams same input CNN (Fig.\ref{fig:5_5streams_same}). The results indicate that the distribution for a 5-streams Streaming Net is closer to the uniform distribution than all other distributions.

\textbf{Distribution of the filter weights in the first conv layer across each stream}.
Here we compare filter weight distribution in the first conv layer of a 1-stream simple CNN vs. each stream of a 5-stream network. The results are presented in Fig. \ref{fig:euro_weight_distrib_5streams}. In this particular case, filter weights distribution of each particular stream of 5-stream Streaming Network is more similar to the unfirm distribution than the distribution for a 1-stream CNN

In general, for Streaming Networks filter weights distribution for a single stream can be more different from the uniform than the distribution for a 1-stream CNN, however, the overall distribution of weights within the layer is more similar to a uniform and this similarity increases as we increase the number of streams. We provide more examples in the Supplementary materials.

\section{Discussion and Conclusion}

Our results indicate that Streaming Net is the only network architecture that enables noise robustness. Furthermore, for the higher capacity models, noise robustness appears only in the case when both hard-wired and input induced-sparsity are employed. However, architectures employing either type of sparsity exhibit no noise robustness. This is summarized in Table \ref{tab:results}.

Regarding filter weights distributions, we have illustrated that filter weights diversity increases and distribution in the first conv layer approaches uniform distribution as hard-wired (number of streams) and input-induced (number of slices) sparsity and capacity grow.

We also speculate that robust recognition of noisy data (previously unseen data) is achieved by a greater diversity of filters created by interplay of two types of spacity. 

To summarize, through a series of tests, we found that hard-wired and input-induced sparsity taken together enable noise robustness and introduce Streaming Nets as a new simple method for robust recognition of noisy images.

{\small

}

\appendix

\subsection{Addinitional Expermental Results}

For our experiments we use three datasets. The selected datasets are cifar10\footnote{https://www.cs.toronto.edu/~kriz/cifar.html}, Eurosat (rgb)\footnote{https://github.com/phelber/eurosat} \cite{Helber2017EuroSAT} and UCmerced land use\footnote{http://weegee.vision.ucmerced.edu/datasets/landuse.html}.

\subsection{Cifar10}

Cifar10 dataset contains RGB 32x32 images of 10 classes (airplane, automobile, bird, cat, deer, dog, frog, horse, ship, truck). The total number of images is 60,000 with 6,000 images for each class. To train and test the networks, we use 50,000 and 10,000 images respectively.

The results of performance comparison are presented as follows: 

- 1-stream simple conv net vs. 5-stream Streaming Net (Fig. \ref{fig:res_cifar10_0001});

\begin{figure*}[t]
\begin{center}
\includegraphics[width=1.0\linewidth]{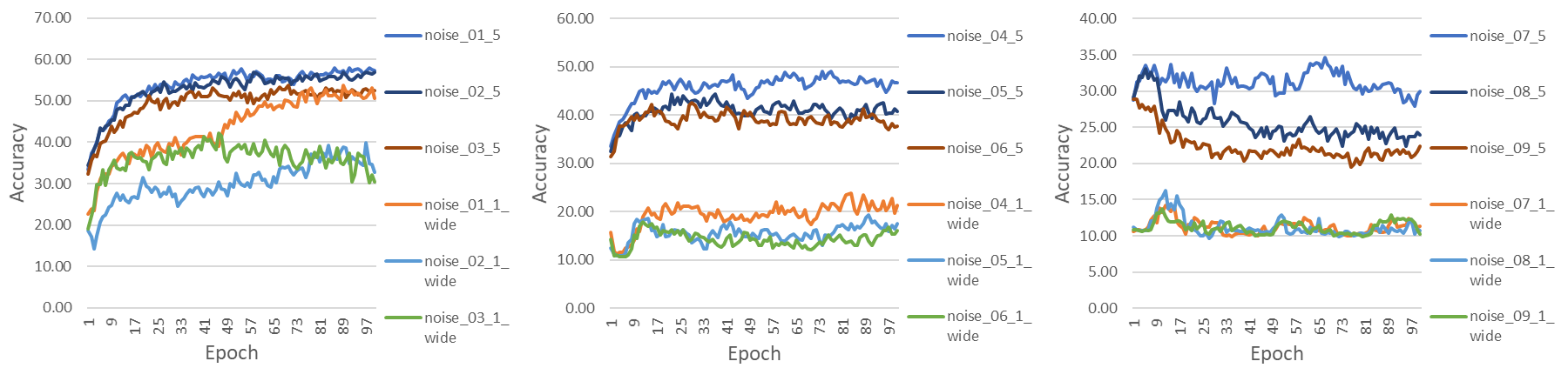}
\end{center}
\caption{Tests of Cifar10 dataset using Adam optimizer with 0.0001 learning rate for 1-stream wide conv net  vs. 5-stream Streaming Net. The notations are the same as in Fig. \ref{fig:res_cifar10_0001}.}
\label{fig:res_cifar10_1str_wide}
\end{figure*}

- 1-stream wide conv net vs. 5-stream Streaming Net (Fig. \ref{fig:res_cifar10_1str_wide});

\begin{figure*}[t]
\begin{center}
\includegraphics[width=1.0\linewidth]{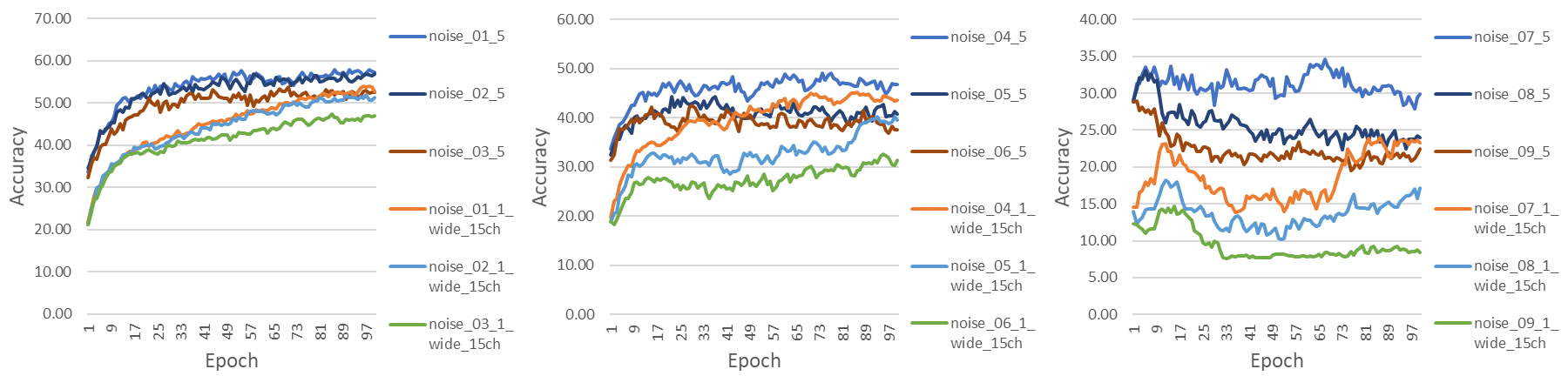}
\end{center}
\caption{Tests of Cifar10 dataset using Adam optimizer with 0.0001 learning rate for 1-stream slice 15-channel wide conv net  vs. 5-stream Streaming Net. The notations are the same as in Fig. \ref{fig:res_cifar10_0001}.}
\label{fig:res_cifar10_15ch_wide}
\end{figure*}

- 1-stream slice 15-channel wide conv net vs. 5-stream Streaming Net (Fig. \ref{fig:res_cifar10_15ch_wide});

\begin{figure*}[t]
\begin{center}
\includegraphics[width=1.0\linewidth]{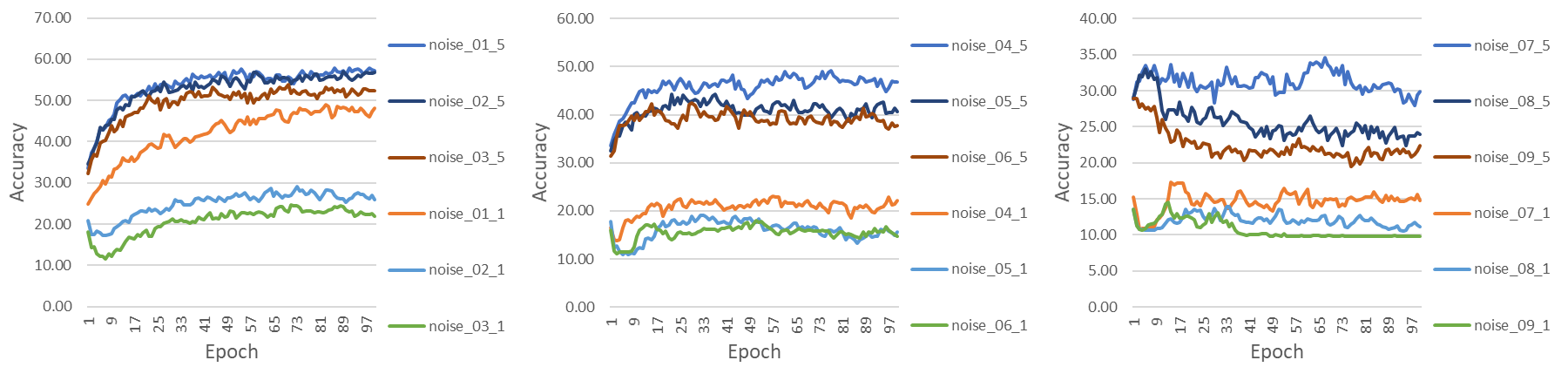}
\end{center}
\caption{Tests of Cifar10 dataset using Adam optimizer with 0.0001 learning rate for 5-stream the same inputs conv net vs. 5-stream Streaming Net. The notations are the same as in Fig. \ref{fig:res_cifar10_0001}.}
\label{fig:res_cifar10_5same}
\end{figure*}

- 5-stream the same inputs conv net vs. 5-stream Streaming Net (Fig. \ref{fig:res_cifar10_5same}).


\subsection{Eurosat}

Eurosat dataset contains Sentinel-2 satellite images covering both 13 spectral bands and RGB (3-channel) 64x64 images and consisting of 10 classes (AnnualCrop, Forest, Herbaceous, Vegetation, Highway, Industrial, Pasture, PermanentCrop, Residential, River, SeaLake) with in total 27,000 labeled and geo-referenced images. We use only RGB images for this study.

The results of performance comparison are presented as follows: 

- 1-stream simple conv net vs. 5-stream Streaming Net (Fig. \ref{fig:res_euro_0001});

- 1-stream wide conv net vs. 5-stream Streaming Net Fig. (Fig. \ref{fig:res_euro_1stream_wide});

- 1-stream slice 15-channel wide conv net vs. 5-stream Streaming Net (Fig. \ref{fig:res_euro_15ch_5_0001});

- 5-stream the same inputs conv net vs. 5-stream Streaming Net (Fig. \ref{fig:res_euro_5same});

- 5-stream vs. 10-stream Streaming Net (Fig. \ref{fig:res_euro_5x10_0001}).


\begin{figure*}[t]
\begin{center}
\includegraphics[width=1.0\linewidth]{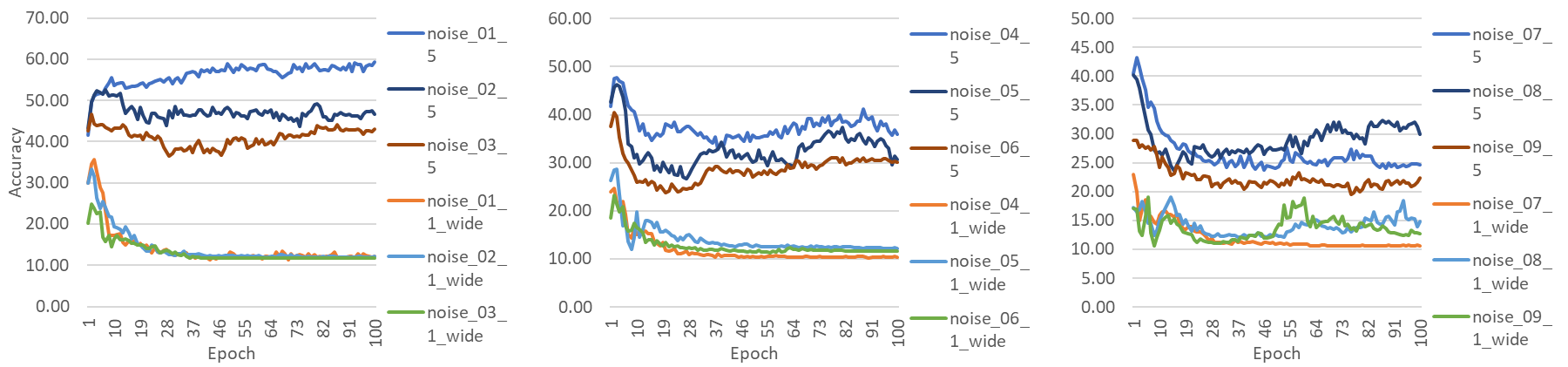}\end{center}
\caption{Tests of Eurosat dataset using Adam optimizer with 0.0001 learning rate for 1-stream wide conv net vs. 5-stream Streaming Net. The notations are the same as in Fig. \ref{fig:res_cifar10_0001}.}
\label{fig:res_euro_1stream_wide}
\end{figure*}

\begin{figure*}[t]
\begin{center}
\includegraphics[width=1.0\linewidth]{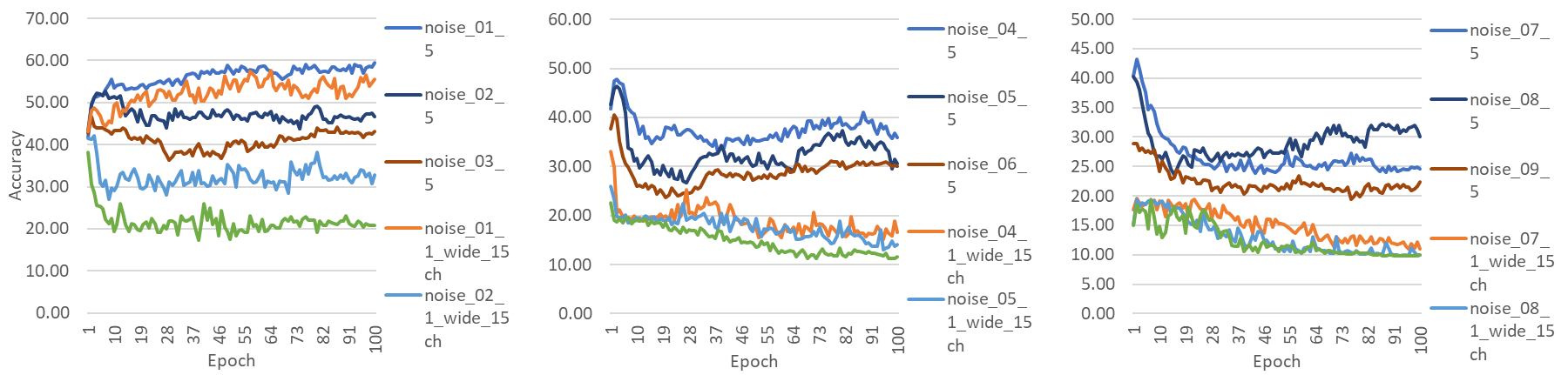}\end{center}
\caption{Tests of Eurosat dataset using Adam optimizer with 0.0001 learning rate for 1-stream 15-channel wide conv net vs. 5-stream Streaming Net. The notations are the same as in Fig. \ref{fig:res_cifar10_0001}.}
\label{fig:res_euro_15ch_5_0001}
\end{figure*}

\begin{figure*}[t]
\begin{center}
\includegraphics[width=1.0\linewidth]{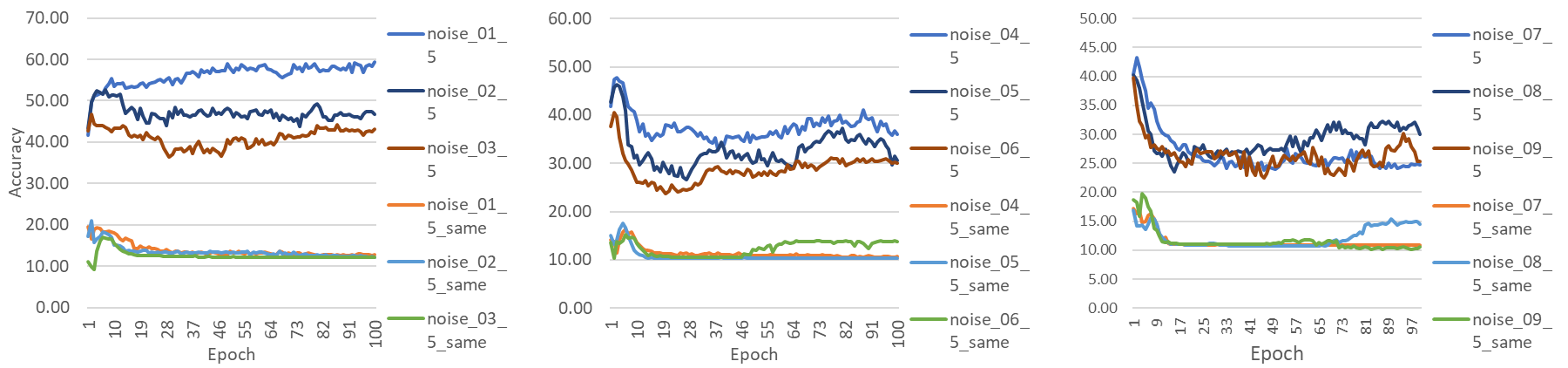}\end{center}
\caption{Tests of Eurosat dataset using Adam optimizer with 0.0001 learning rate for 5-streams with the same inputs conv net vs. 5-stream Streaming Net. The notations are the same as in Fig. \ref{fig:res_cifar10_0001}.}
\label{fig:res_euro_5same}
\end{figure*}

\begin{figure*}[t]
\begin{center}
\includegraphics[width=1.0\linewidth]{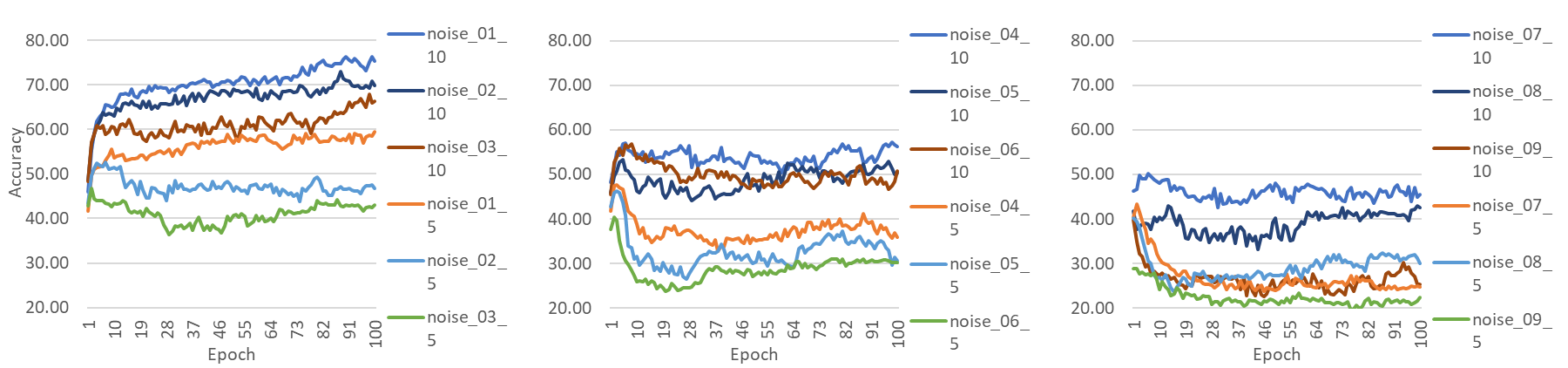}\end{center}
\caption{Tests of Eurosat dataset using Adam optimizer with 0.0001 learning rate for 5-streams vs. 10-stream Streaming Net. The notations is the same as in Fig. \ref{fig:res_cifar10_0001}.}
\label{fig:res_euro_5x10_0001}
\end{figure*}

\subsection{UCmerced}

The UCmerced land-use dataset contains 256x256 RGB satellite images of 21 class of land use. There are 100 images for each of the following classes: agricultural, airplane, baseballdiamond, beach, buildings, chaparral, denseresidential, forest, freeway, golfcourse, harbor, intersection, mediumresidential, mobilehomepark, overpass, parkinglot, river, runway, sparseresidential, storagetanks, tenniscourt.

The results of performance comparison are presented as follows: 

- 1-stream simple conv net vs. 5-stream Streaming Net (Fig. \ref{fig:res_ucmerced_0001});

\begin{figure*}[t]
\begin{center}
\includegraphics[width=1.0\linewidth]{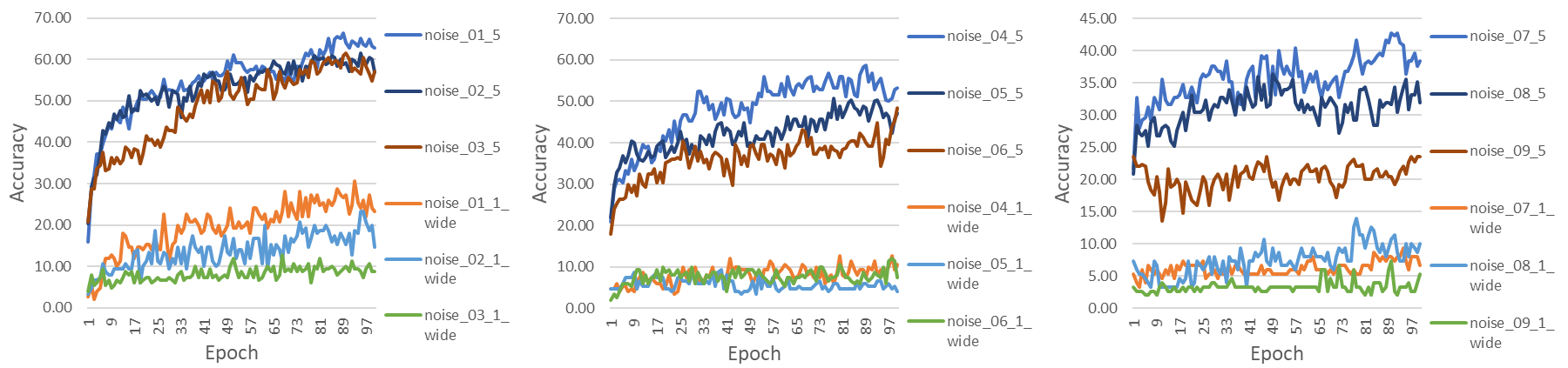}
\end{center}
\caption{Tests of UCmerced dataset using Adam optimizer with 0.0001 learning rate for 1-stream wide conv net vs. 5-stream Streaming Network. The notations are the same as in Fig. \ref{fig:res_cifar10_0001}}
\label{fig:res_ucmerced_0001_1stream_wide}
\end{figure*}

- 1-stream wide conv net vs. 5-stream Streaming Net Fig. (Fig. \ref{fig:res_ucmerced_0001_1stream_wide});

\begin{figure*}[t]
\begin{center}
\includegraphics[width=1.0\linewidth]{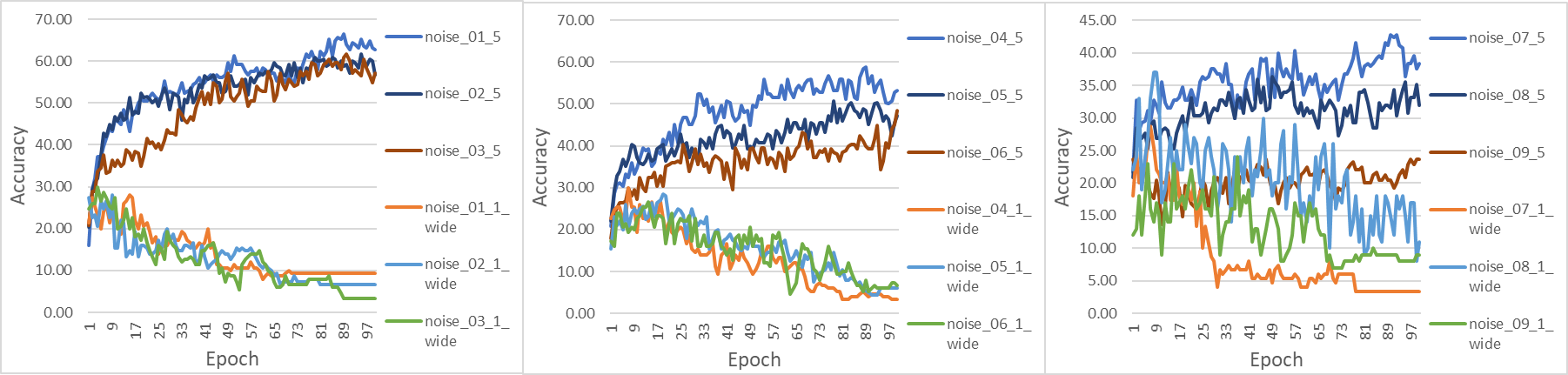}
\end{center}
\caption{Tests of UCmerced dataset using Adam optimizer with 0.0001 learning rate for 1-stream slice 15-channel wide conv net vs. 5-stream Streaming Network. The notations are the same as in Fig. \ref{fig:res_cifar10_0001}}
\label{fig:res_ucmerced_0001_1_15ch_wide}
\end{figure*}

- 1-stream slice 15-channel wide conv net vs. 5-stream Streaming Net (Fig. \ref{fig:res_ucmerced_0001_1_15ch_wide});

\begin{figure*}[t]
\begin{center}
\includegraphics[width=1.0\linewidth]{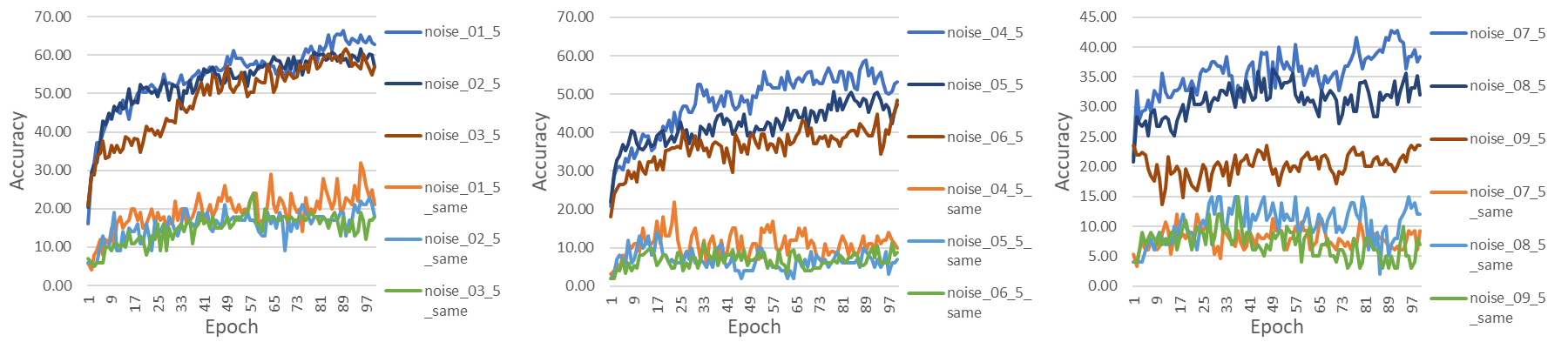}
\end{center}
\caption{Tests of UCmerced dataset using Adam optimizer with 0.0001 learning rate for 5-stream with the same input conv net vs. 5-stream Streaming Network. The notations are the same as in Fig. \ref{fig:res_cifar10_0001}}
\label{fig:res_ucmerced_0001_5streams_same}
\end{figure*}

- 5-stream the same inputs conv net vs. 5-stream Streaming Net (Fig. \ref{fig:res_ucmerced_0001_5streams_same}).


\subsection{Addinitional Experimental Results: Filter Weights Distributions for the first conv layer using Eurosat dataset}

In this section, we introduce the results of the analysis of distributions of filter weights after training neural networks using Eurosat dataset. We introduce results for both overall filter weights distributions and stream-to-stream distributions comparison. Overall filter weights distributions imply that all the filter weights are collected from all the streams and filters of the first conv layers. Stream-to-stream distributions comparison means that we compare all the filter weights from all the filters in the first conv layer of a 1-stream simple conv net vs. all the filter weights from all the filters in the first conv layer of for each stream of the corresponding Streaming Network.

\textbf{Distribution of the filter weights across individual conv layers}

Here we discuss distribution of filter weights for a 1-stream simple conv net, 5- and 10-stream Streaming networks in the case of Eurosat data set.

\begin{figure*}[t]
\begin{center}
\includegraphics[width=1.0\linewidth]{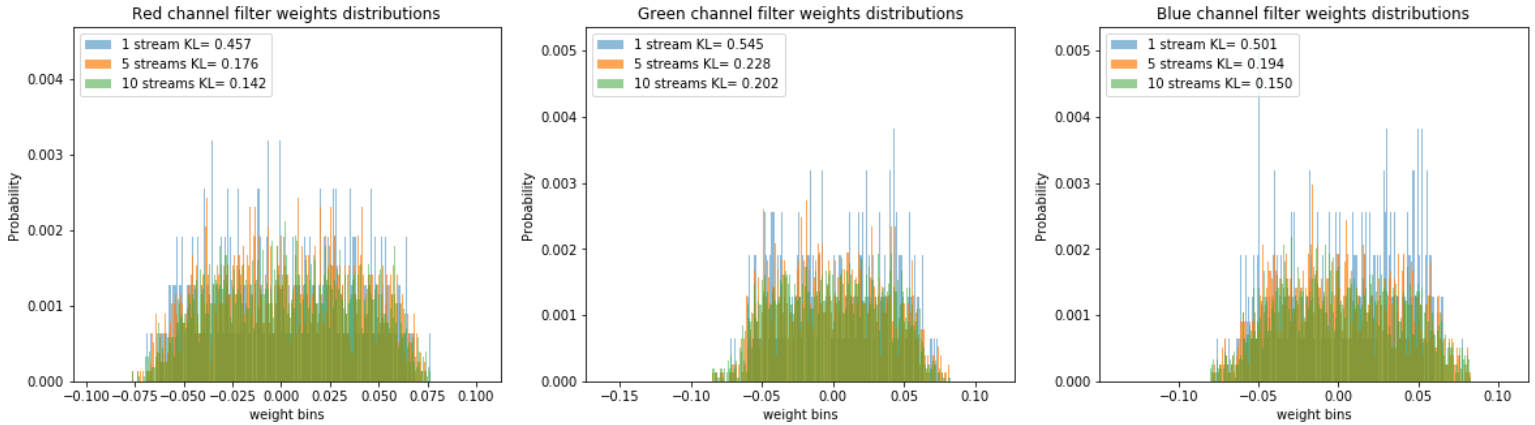}
\end{center}
\caption{Test 1: the overall distribution of the filter weights across all the layers and filters of the first conv layers for Eurosat dataset. Notation: ``1 stream" implies 1-stream simple conv net, ``5 streams" implies 5-stream Streaming Network, ``10 streams" implies 10-stream Streaming Network. KL is a short for Kullback–Leibler divergence.}
\label{fig:euro_weight_distrib_all_streams_1}
\end{figure*}

\begin{figure*}[t]
\begin{center}
\includegraphics[width=1.0\linewidth]{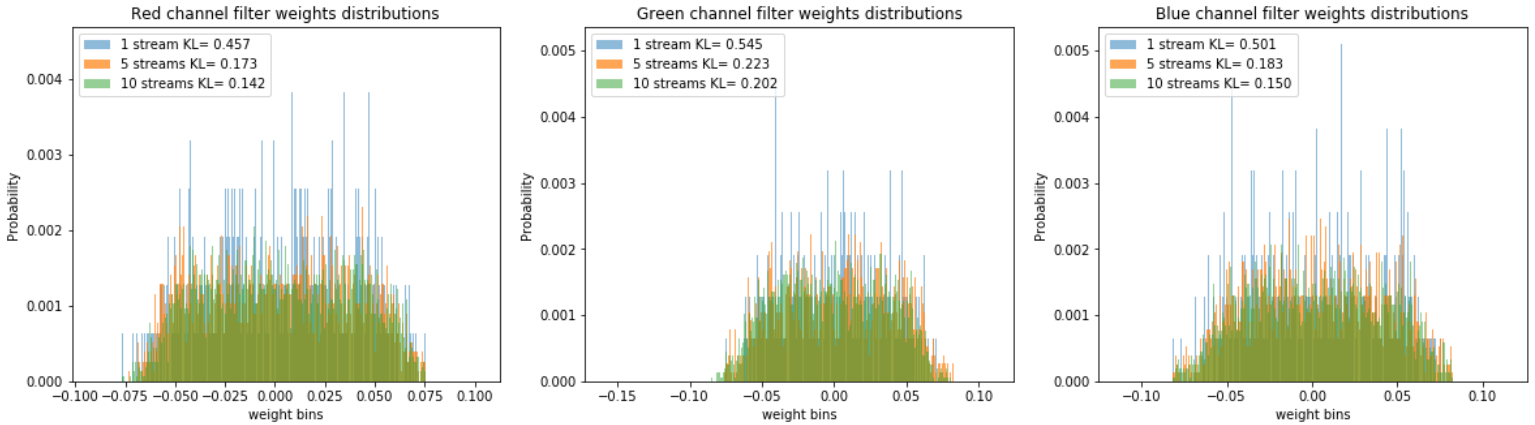}
\end{center}
\caption{Test 2: the overall distribution of the filter weights across all the layers and filters of the first conv layers for Eurosat dataset. Notations are the same as in Fig. \ref{fig:euro_weight_distrib_all_streams_1}.}
\label{fig:euro_weight_distrib_all_streams_2}
\end{figure*}

\begin{figure*}[t]
\begin{center}
\includegraphics[width=1.0\linewidth]{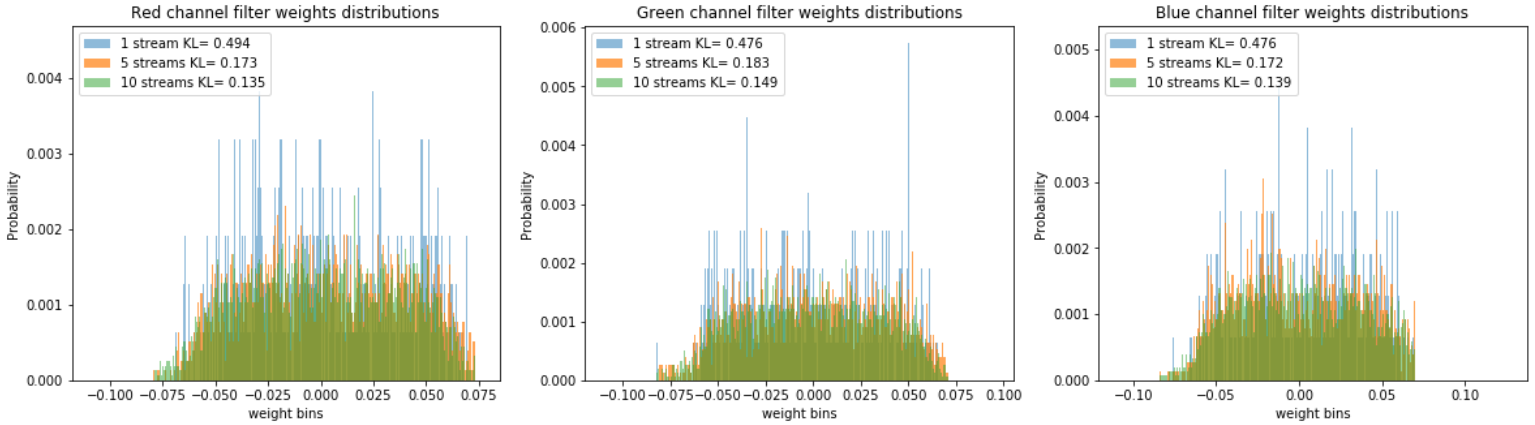}
\end{center}
\caption{Test 3: the overall distribution of the filter weights across all the layers and filters of the first conv layers for Eurosat dataset. Notations are the same as in Fig. \ref{fig:euro_weight_distrib_all_streams_1}.}
\label{fig:euro_weight_distrib_all_streams_3}
\end{figure*}

\begin{figure*}[t]
\begin{center}
\includegraphics[width=1.0\linewidth]{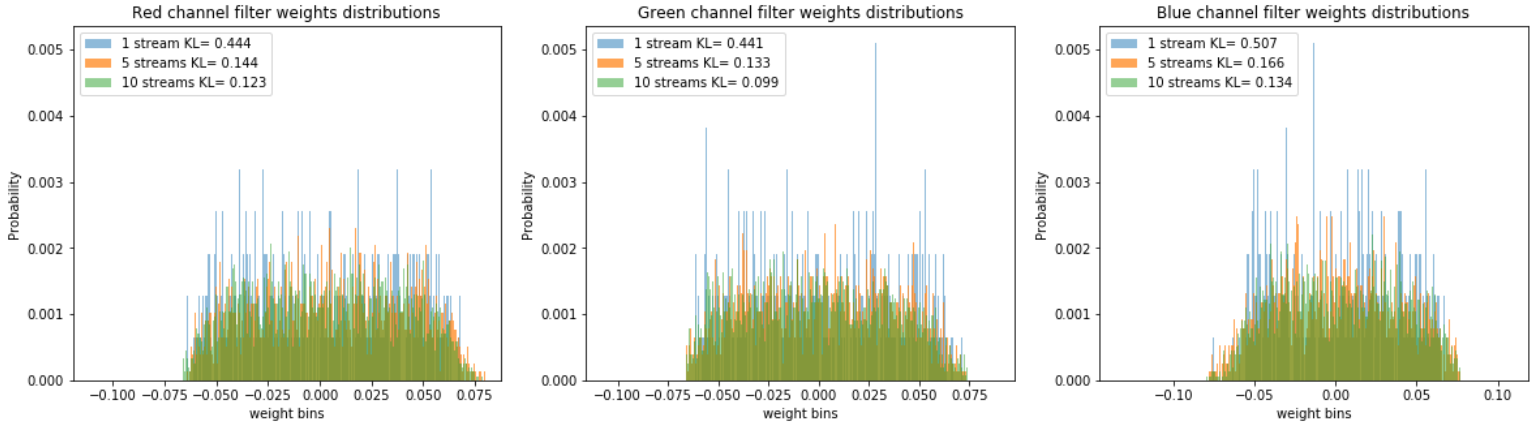}
\end{center}
\caption{Test 4: the overall distribution of the filter weights across all the layers and filters of the first conv layers for Eurosat dataset. Notations are the same as in Fig. \ref{fig:euro_weight_distrib_all_streams_1}.}
\label{fig:euro_weight_distrib_all_streams_4}
\end{figure*}

\begin{figure*}[t]
\begin{center}
\includegraphics[width=1.0\linewidth]{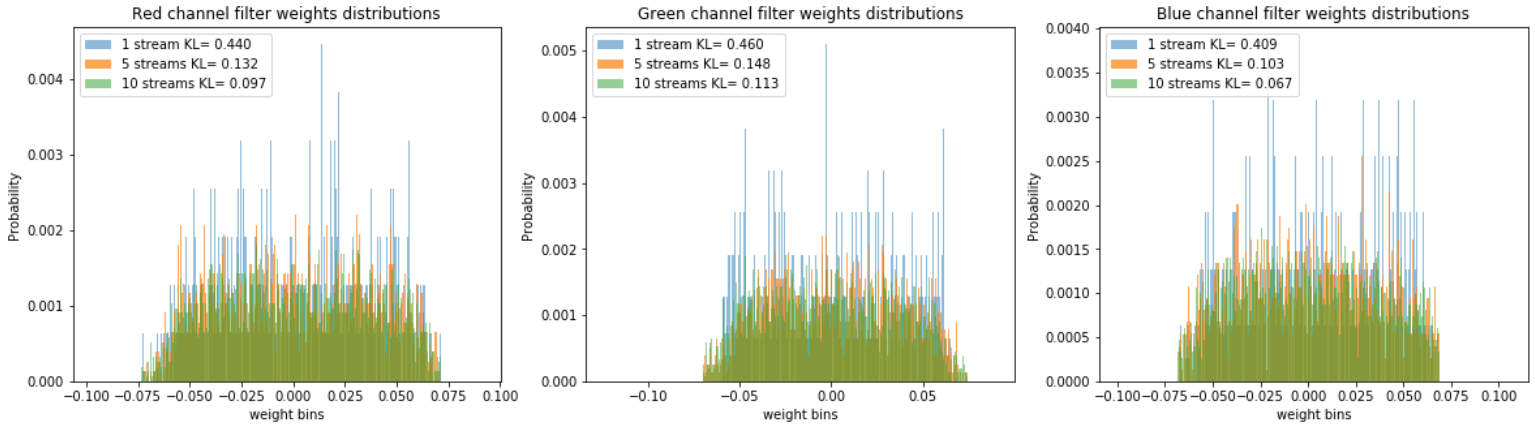}
\end{center}
\caption{Test 5: the overall distribution of the filter weights across all the layers and filters of the first conv layers for Eurosat dataset. Notations are the same as in Fig. \ref{fig:euro_weight_distrib_all_streams_1}.}
\label{fig:euro_weight_distrib_all_streams_5}
\end{figure*}

\begin{figure*}[t]
\begin{center}
\includegraphics[width=1.0\linewidth]{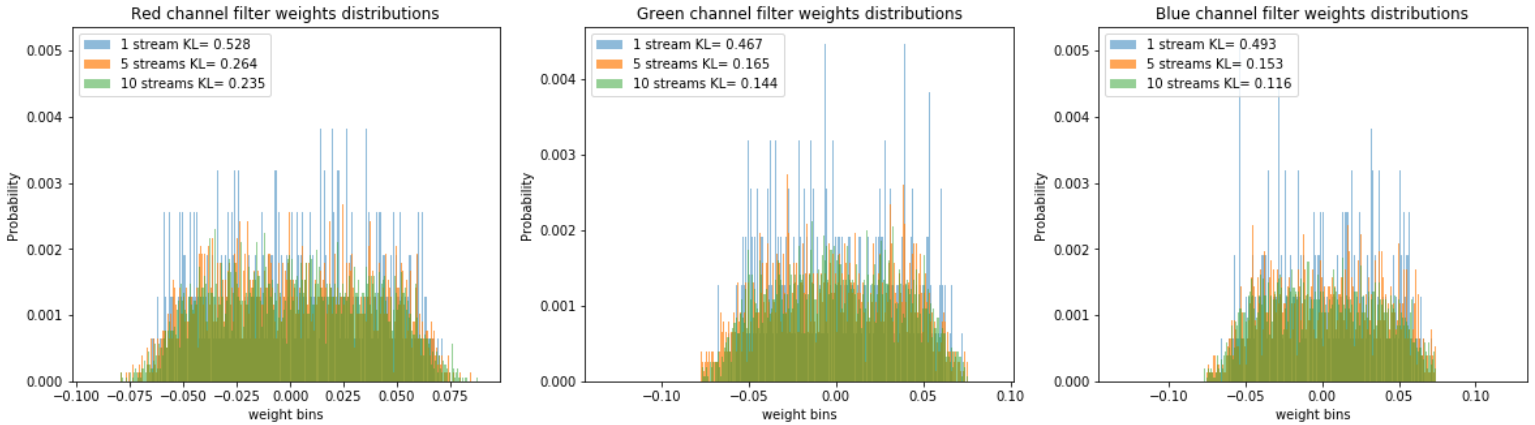}
\end{center}
\caption{Test 6: the overall distribution of the filter weights across all the layers and filters of the first conv layers for Eurosat dataset. Notations are the same as in Fig. \ref{fig:euro_weight_distrib_all_streams_1}.}
\label{fig:euro_weight_distrib_all_streams_6}
\end{figure*}

\begin{figure*}[t]
\begin{center}
\includegraphics[width=1.0\linewidth]{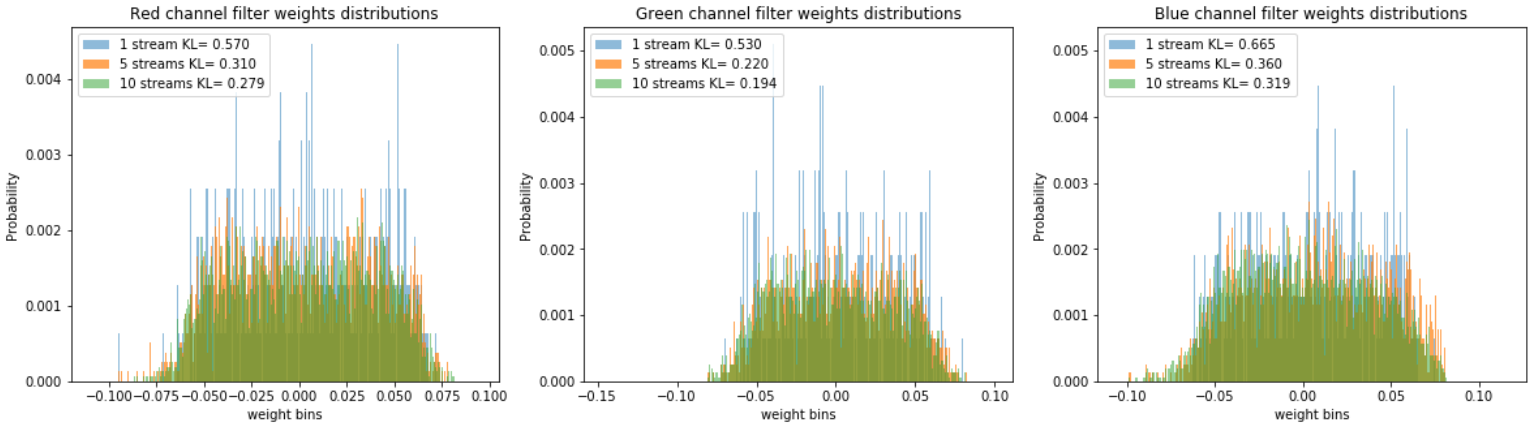}
\end{center}
\caption{Test 7: the overall distribution of the filter weights across all the layers and filters of the first conv layers for Eurosat dataset. Notations are the same as in Fig. \ref{fig:euro_weight_distrib_all_streams_1}.}
\label{fig:euro_weight_distrib_all_streams_7}
\end{figure*}

\begin{figure*}[t]
\begin{center}
\includegraphics[width=1.0\linewidth]{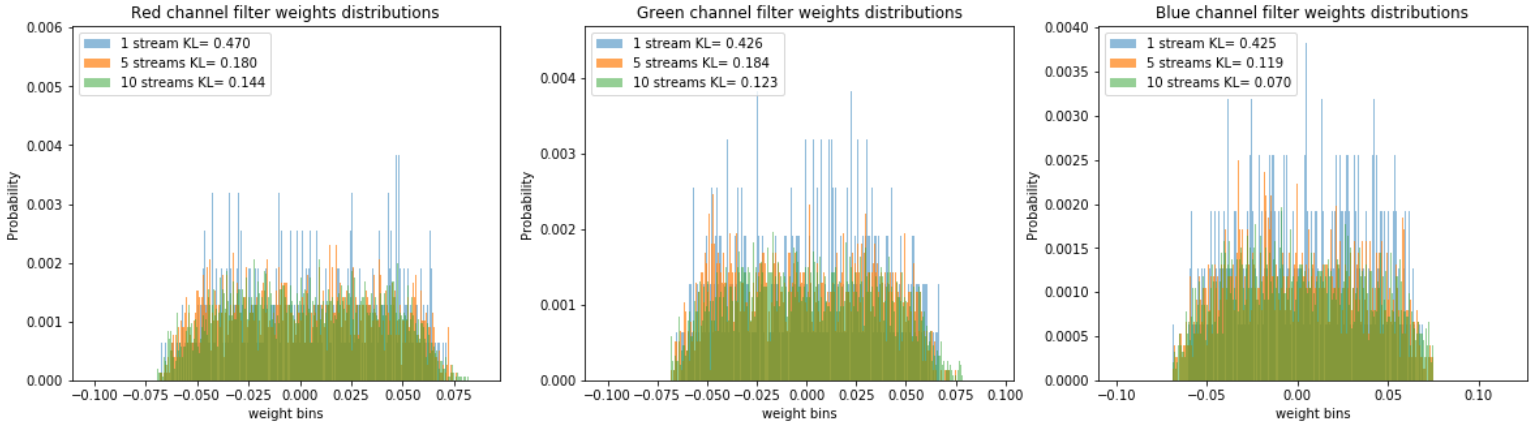}
\end{center}
\caption{Test 8: the overall distribution of the filter weights across all the layers and filters of the first conv layers for Eurosat dataset. Notations are the same as in Fig. \ref{fig:euro_weight_distrib_all_streams_1}.}
\label{fig:euro_weight_distrib_all_streams_8}
\end{figure*}

We performed an analysis of 8 different sets of weights for 1-stream simple conv net, 5- and 10-stream Streaming Net. We illustrate the distribution of filter weight for the first conv layer of the 1-stream conv net and 5- and 10-stream Streaming Network separately for each color channel for each of 8 triplets in Figs. \ref{fig:euro_weight_distrib_all_streams_1} through \ref{fig:euro_weight_distrib_all_streams_8}. In each figures (a), (b) and (c) sections introduce distributions of weights for red, green and blue channels, respectively.

One can infer that KL divergence value decreases as we increase the number of streams. Therefore, the distribution of filter weights is gradually approaching a discrete uniform distribution.

\begin{figure*}[t]
\begin{center}
\includegraphics[width=1.0\linewidth]{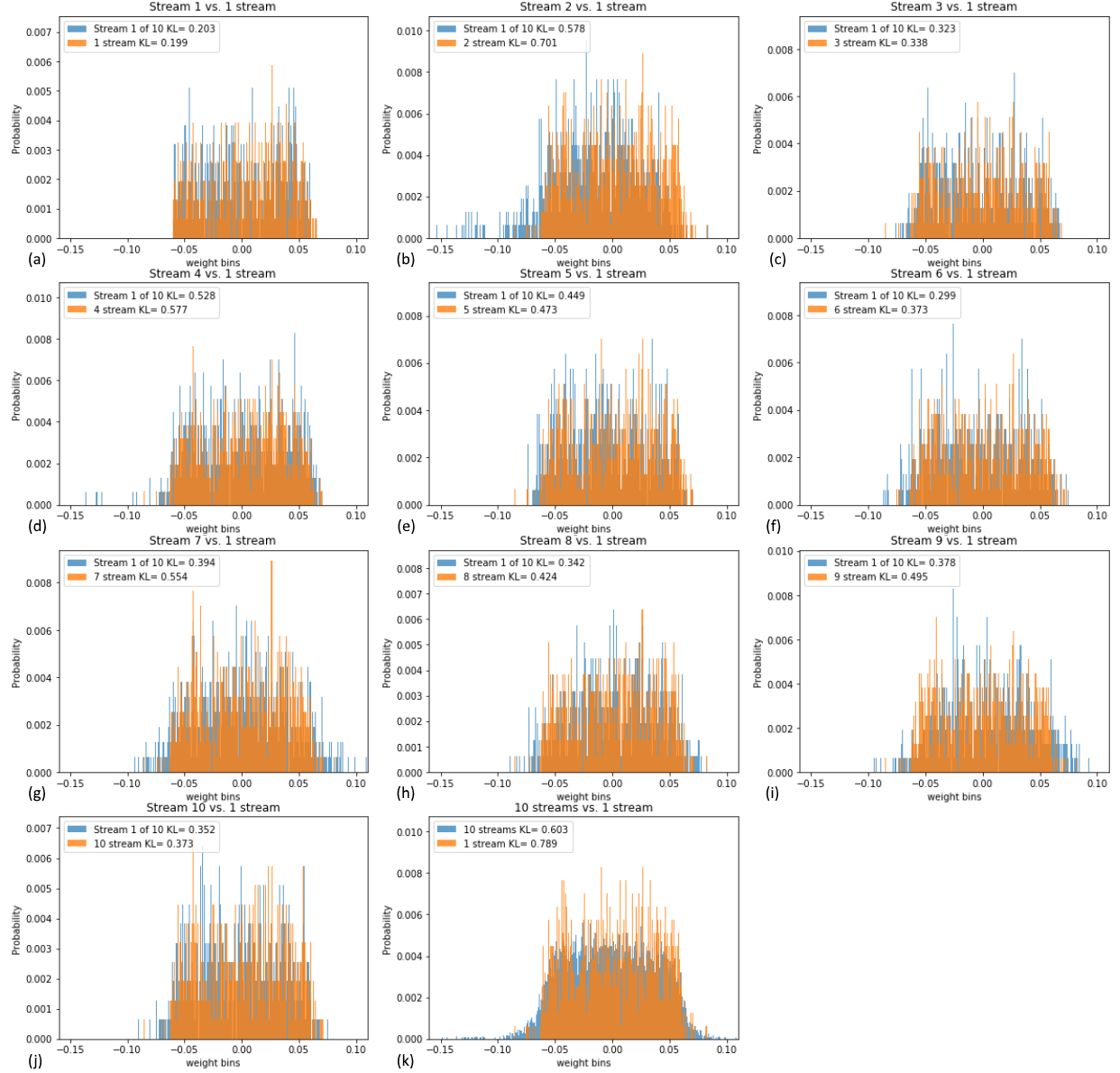}
\end{center}
\caption{Distribution of filter weights across all layers for each stream of 10-stream Streaming Net vs. a 1-stream simple conv net obtained for Eurosat dataset.}
\label{fig:euro_weight_distrib_10streams}
\end{figure*}

\textbf{Distribution of the filter weights in the first conv layer across each stream}
Here we compare filter weight distribution in the first conv layer of a 1-stream simple conv net, each stream of 5- and 10-stream Streaming Network. The results of the comparison between 1-stream simple conv net and 5-stream Streaming Net are presented in Fig. \ref{fig:euro_weight_distrib_5streams}. The results of the comparison between 1-stream simple conv net and 10-stream Streaming Net are presented in Fig. \ref{fig:euro_weight_distrib_10streams}. In Figs. \ref{fig:euro_weight_distrib_5streams} and \ref{fig:euro_weight_distrib_10streams}, we introduce a stream-to-stream comparison only for red channel for one of the tests.

Figs. \ref{fig:euro_weight_distrib_5streams}(a)-(e) and Figs. \ref{fig:euro_weight_distrib_10streams}(a)-(j) introduce stream-to-stream comparison of filter weights distributinos between each stream of 5-stream Streaming Net and a 1-stream simple conv net, and 10-stream Streaming Net and a 1-stream simple conv net, respectively.

In general, filter weights distribution of each particular stream of 5-stream Streaming Network is more similar to the uniform distribution than the distribution for a 1-stream conv net. However, for 10-stream Streaming Net there are cases when filter weights distribution of particular single stream is more different from the uniform distribution than the distribution for 1-stream network.

Nevertheless, the overall weight distribution across all the streams for both 5- and 10-stream Streaming Nets are more similar to the uniform disctrition than a corresponding disctribution derived from 1-stream simple conv net as shown in Fig. \ref{fig:euro_weight_distrib_5streams}(f) and Fig. \ref{fig:euro_weight_distrib_10streams}(k), respectively.

\begin{figure*}[t]
\begin{center}
\includegraphics[width=1.0\linewidth]{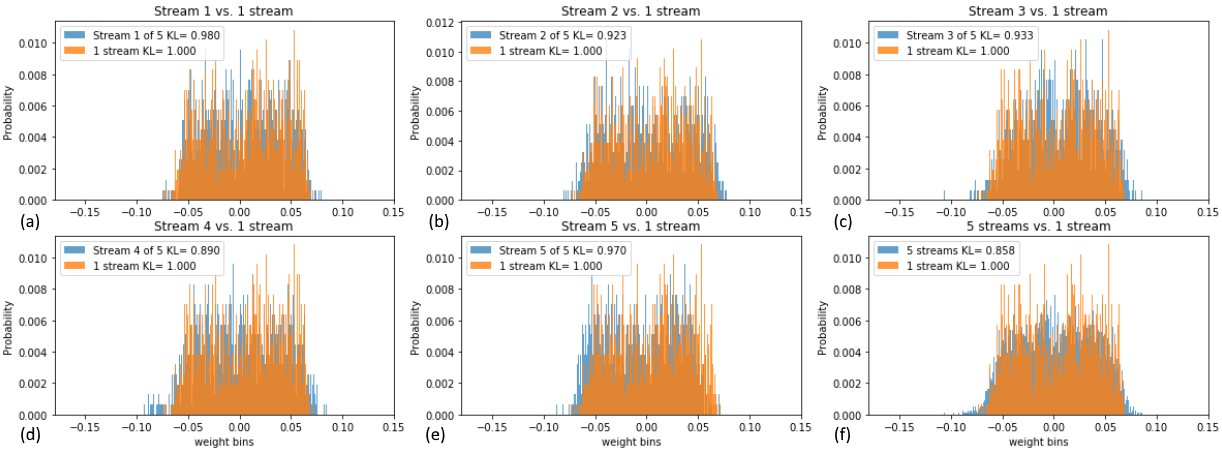}
\end{center}
\caption{Distribution of filter weights across all layers for each stream of 5-stream Streaming Net vs. a 1-stream simple conv net obtained for Eurosat dataset.}
\label{fig:euro_weight_distrib_5streams}
\end{figure*}

\end{document}